\documentclass{article}

\PassOptionsToPackage{numbers, compress}{natbib}

\usepackage[final]{neurips_2025}

\usepackage[utf8]{inputenc} % allow utf-8 input
\usepackage[T1]{fontenc}    % use 8-bit T1 fonts
\usepackage{hyperref}       % hypertclinks
\usepackage{url}            % simple URL typesetting
\usepackage{booktabs}       % professional-quality tables
\usepackage{amsfonts}       % blackboard math symbols
\usepackage{nicefrac}       % compact symbols for 1/2, etc.
\usepackage{microtype}      % microtypography
\usepackage{xcolor}         % colors

\usepackage{tikz}
\usepackage{adjustbox}
\usepackage{amsmath, amssymb, bm}
\usepackage{xcolor}
\usepackage{pgfplots}
\usepackage{caption}
\usepackage{subcaption}
\usepackage{xspace}
\usepackage[most]{tcolorbox}

\usepackage{graphicx}
\usepackage{booktabs}
\usepackage{multirow}
\usepackage{multicol}
\usepackage{adjustbox}
\usepackage{enumitem}

\def\eqref#1{equation~\ref{#1}}
\def\1{\bm{1}}

\DeclareMathAlphabet{\mathsfit}{\encodingdefault}{\sfdefault}{m}{sl}
\SetMathAlphabet{\mathsfit}{bold}{\encodingdefault}{\sfdefault}{bx}{n}

\newcommand{\softmax}{\mathrm{softmax}}

\newcommand{\concat}{%
  \mathbin{%
    \text{\textcircled{\scriptsize$\boldsymbol{;}$}}%
  }%
}

\newcommand{\methodname}{\textsc{wilda}\xspace}
\newcommand{\wildas}{\textsc{wilda-s}\xspace}
\newcommand{\wildaf}{\textsc{wilda-f}\xspace}
\newcommand{\wildar}{\textsc{wilda-r}\xspace}
\newcommand{\sst}{\textsc{sst}\xspace}
\newcommand{\cola}{\textsc{cola}\xspace}
\newcommand{\rte}{\textsc{rte}\xspace}
\newcommand{\mrpc}{\textsc{mrpc}\xspace}
\newcommand{\qqp}{\textsc{qqp}\xspace}
\newcommand{\qnli}{\textsc{qnli}\xspace}
\newcommand{\mnli}{\textsc{mnli}\xspace}
\newcommand{\elmath}{\textsc{math}\xspace}
\newcommand{\misc}{\textsc{misc}\xspace}

\newenvironment{example}[1][Example]{%
    \vspace{1em}
    \noindent
    \begin{tcolorbox}[colback=lightgray!5!white, colframe=lightgray!75!black, breakable, title=#1] % Use green colors and make it breakable
}{%
    \end{tcolorbox}
}

\usetikzlibrary{shapes, arrows.meta, positioning, decorations.pathreplacing, calc}
\definecolor{salmonpink}{RGB}{255,145,164}
\definecolor{turquoise}{RGB}{64,224,208}
\definecolor{lavender}{RGB}{220,208,255}
\definecolor{myblue}{RGB}{21,101,192}
\definecolor{myred}{RGB}{198,40,40}
\definecolor{mygreen}{RGB}{35,101,51}

\title{Disentangling Latent Shifts of In-Context Learning with Weak Supervision}

\author{%
 Josip Juki{\'{c}}, \ Jan {\v{S}}najder\\
TakeLab \\ Faculty of Electrical Engineering and Computing \\ University of Zagreb, Croatia\\
\tt \{josip.jukic, jan.snajder\}@fer.hr
}

\begin{document}

\maketitle

\begin{abstract}

In-context learning (ICL) enables large language models to perform few-shot learning by conditioning on labeled examples in the prompt. Despite its flexibility, ICL suffers from instability -- especially as prompt length increases with more demonstrations. To address this, we treat ICL as a source of weak supervision and propose a parameter-efficient method that disentangles demonstration-induced latent shifts from those of the query. An ICL-based teacher generates pseudo-labels on unlabeled queries, while a student predicts them using only the query input, updating a lightweight adapter. This captures demonstration effects in a compact, reusable form, enabling efficient inference while remaining composable with new demonstrations. Although trained on noisy teacher outputs, the student often outperforms its teacher through pseudo-label correction and coverage expansion, consistent with the weak-to-strong generalization effect. Empirically, our method improves generalization, stability, and efficiency across both in-domain and out-of-domain tasks, surpassing standard ICL and prior disentanglement methods.

\end{abstract}

\section{Introduction}

% ICL intro
In-context learning (ICL) has become a core mechanism for adapting large language models (LLMs) to new tasks in a way that removes the need to update their parameters \citep{brown-etal-2020-language, dong-etal-2024-survey}. By prepending a few labeled examples, called \textit{demonstrations}, to the input query, LLMs can perform few-shot learning directly at inference time. This paradigm is especially attractive in low-resource settings, where full fine-tuning is too costly or impractical.

% ICL suffers from instability, inefficiency, and limited scalability 
Despite its convenience, ICL performance is highly sensitive to the selection and ordering of demonstrations, often resulting in unstable predictions and poor generalization \citep{lu-etal-2021-fantastically, li-etal-2024-debiasing}. Moreover, ICL typically requires long contexts, as multiple demonstrations must be included alongside the query in a single input. As input lengths grow, inference becomes increasingly inefficient, inflating processing costs, amplifying positional biases -- including inherent primacy and recency effects in transformer-based LLMs \citep{liu-etal-2024-lost} -- and pushing against the model’s context window limits \citep{dong-etal-2024-exploring}. Consequently, ICL scales poorly with the number of demonstrations \citep{chen-etal-2023-many, cai-etal-2023-scaling}: beyond a certain threshold, additional examples either degrade performance or must be discarded entirely. This inefficiency and poor scalability limit ICL’s ability to incorporate more supervision, preventing it from fully leveraging the potential benefits of richer demonstrations.

% Disentangling latent shifts offers a path to stability and scalability
To address these limitations, a mechanistic perspective on ICL has proven useful. In this view, demonstrations influence model behavior by inducing \textit{latent shifts} -- context-dependent changes in internal representations that alter how the model processes the query.
Disentangling these shifts from the representation of the query itself allows ICL to operate more robustly, processing queries independently of demonstrations. This, in turn, enables contextual knowledge to be stored persistently, eliminating the need to reprocess demonstrations for each new query. The latent shifts can then be reapplied directly, thereby reducing prompt length, improving inference efficiency, and enabling more modular and reusable representations.
The disentanglement of demonstration-induced latent shifts from those of the query has been explored from both practical and theoretical perspectives: some work aims to improve the stability and scalability of ICL \citep{liu-etal-2023-context, zhang-etal-2024-batch}, while other studies provide formal insights into the nature of context-induced shifts \citep{dai-etal-2023-gpt, todd-etal-2024-function}.
However, existing methods rely on approximations that intervene directly in attention heads or hidden states. In contrast, we adopt a functional perspective on ICL, capturing demonstration-induced shifts implicitly through model outputs rather than architectural manipulation. This enables disentanglement to emerge as a learned, self-aligned process rather than one achieved through explicit manipulation of internal states.

% ICL as a source of weak supervision
In this paper, we propose to disentangle the latent shifts of the demonstrations and the query by using the output of ICL as a weak supervision signal, rather than approximating the underlying internal mechanisms. Specifically, we use ICL predictions as pseudo-labels that capture the full, contextualized influence of demonstrations. These pseudo-labels guide the training of a student model that internalizes demonstration-conditioned behavior -- without repeated prompting or architectural intervention.
We instantiate this idea with \methodname{} (\textbf{W}eakly-supervised \textbf{I}n-context \textbf{L}earning \textbf{D}isentanglement via \textbf{A}dapters), a parameter-efficient method for encoding the latent shifts induced by in-context demonstrations. In a teacher--student setup, an ICL-based teacher generates pseudo-labels for unlabeled queries, while the student learns to predict them using only the query input. The student updates a lightweight adapter module \citep{houlsby-etal-2019-parameter}, enabling it to capture the shift in a reusable and modular form. The adapter supports efficient inference without requiring demonstrations in the prompt, and multiple adapters can be combined through simple arithmetic to fuse knowledge across demonstration subsets. At inference time, additional in-context demonstrations can be composed with the adapter’s shift, enabling flexible integration of prompt-based and parameter-based adaptation.

% Weak-to-strong generalization
We evaluate \methodname{} on both in-domain (ID) and out-of-domain (OOD) data, comparing it to standard ICL, prompt-based fine-tuning \citep{schick-schutze-2021-exploiting}, and recent approaches that manipulate architecture or hidden states to disentangle latent shifts. \methodname{} consistently improves generalization, prompt robustness, and inference efficiency, while remaining highly parameter-efficient.
Despite learning from ICL outputs, the student often surpasses its teacher. We experimentally show that this improvement arises from two emergent behaviors: \textit{pseudo-label correction}, where the student refines noisy or inconsistent outputs from the teacher, and \textit{coverage expansion}, where it generalizes beyond the narrow patterns encoded in the demonstrations. Together, these effects enable \textit{weak-to-strong (W2S) generalization} \citep{lang-etal-2024-theoretical}, allowing the model to learn stable task behavior from limited supervision.

% Contribution
Our contribution is threefold:
(1) we propose \methodname{}, a method that encodes ICL-induced behavior into reusable adapters, improving inference efficiency and prompt stability without requiring demonstration prompts;
(2) we show that \methodname{} outperforms traditional ICL and latent shift disentanglement methods on both ID and OOD data, while maintaining parameter efficiency; and 
(3) we demonstrate that multiple adapters can be combined through simple arithmetic operations, enabling scalable composition of demonstration subsets and efficient handling of long-context scenarios.
Together, these results demonstrate that viewing ICL as a form of weak supervision enables parametric generalization that can also flexibly incorporate contextual adaptation. \methodname{} thus offers a scalable and composable framework for stable task adaptation in LLMs.\footnote{Our code is available at \url{https://github.com/josipjukic/wilda}.}

\begin{figure}[]
\centering
\begin{tikzpicture}[
    teacherbox/.style={rectangle, rounded corners, draw, minimum height=0.8cm, minimum width=3.5cm, fill=gray!30, align=center, line width=0.5mm},
    adapter/.style={rectangle, rounded corners, draw, minimum height=0.8cm, minimum width=1cm, align=center, line width=0.4mm, fill=orange!30},
    fadedadapter/.style={rectangle, rounded corners, draw, minimum height=0.8cm, minimum width=1cm, align=center, line width=0.2mm, fill=orange!20, opacity=0.6},
    fadedadapter2/.style={rectangle, rounded corners, draw, minimum height=0.8cm, minimum width=1cm, align=center, line width=0.2mm, fill=orange!10, opacity=0.3},
    demobox/.style={rectangle, rounded corners, draw=myblue!70, minimum height=1.4cm, minimum width=2.8cm, align=center, fill=myblue!15, line width=0.6mm},
    querybox/.style={rectangle, rounded corners, draw=myred!70, minimum height=0.4cm, minimum width=2cm, align=center, fill=myred!15, line width=0.6mm},
    arrow_teacher/.style={-{Triangle}, very thick, draw=gray!90},
    arrow_student/.style={-{Triangle}, very thick, draw=black!90},
    concatenation/.style={inner sep=1pt},
    studentbox/.style={rectangle, rounded corners, draw, minimum height=1.6cm, minimum width=5cm, fill=gray!10, align=center, line width=0.5mm},
    dashedbox/.style={rectangle, rounded corners, draw, minimum height=0.8cm, minimum width=2.5cm, align=center, line width=0.4mm, fill=orange!15}
    ]

    % Teacher LLM box
    \node[teacherbox] (teacherLLM) {\textbf{Teacher LLM}};

    % Student LLM box below Teacher LLM
    \node[studentbox, below=0.2cm of teacherLLM] (studentLLM) {};
    \node[anchor=north west] at ([xshift=0.1cm, yshift=-0.1cm]studentLLM.north west) {\textbf{Student LLM}};

    % Base LLM and Adapter inside Student LLM
    \node[dashedbox, minimum width=2.5cm, anchor=west, xshift=0.2cm, yshift=-0.2cm] at (studentLLM.west) (baseLLM) {Base LLM};
    \node[fadedadapter2, minimum width=1.8cm, anchor=west, xshift=0.3cm, yshift=0.2cm] at (baseLLM.east) {};
    \node[fadedadapter, minimum width=1.8cm, anchor=west, xshift=0.2cm, yshift=0.1cm] at (baseLLM.east) {};
    \node[adapter, minimum width=1.8cm, anchor=west, right=0.1cm of baseLLM] (adapter) {Adapter};

    % Demonstrations and Query nodes centered at the midpoint
    \node[demobox, left=2.5cm of teacherLLM, yshift=-0.3cm] (demonstrations) {Demonstrations:\\[5pt] $\mathbf{X}_d = [\mathbf{x}_1,\, \mathbf{x}_2,\, \ldots,\, \mathbf{x}_n]$};
    \node[querybox, below=0.5cm of demonstrations] (query) {Query: $\mathbf{x}_q$};

    % Concatenation operator (⊕)
    \node[concatenation, right=0.75cm of demonstrations] (concat) {\scalebox{1.5}{$\concat$}};

    % Connect Demonstrations to the concatenation operator
    \draw [arrow_teacher] (demonstrations.east) -- (concat.west);

    % Connect Query to the concatenation operator
    \draw [arrow_teacher] (query.east) to[out=0, in=230] (concat.south west);

    % Connect the concatenation operator to the Teacher LLM
    \draw [arrow_teacher] (concat.east) to[out=0, in=180] (teacherLLM.west);

    % Teacher output yt
    \node[right=1.25cm of teacherLLM] (yt) {$\mathbf{y}_t$};
    \draw [arrow_teacher] (teacherLLM.east) -- (yt.west);

    % Connect Query to Student LLM
    \draw [arrow_student] (query.east) to[out=0, in=180] (studentLLM.west);

    % Student output ys
    \node[right=0.5cm of studentLLM] (ys) {$\mathbf{y}_s$};
    \draw [arrow_student] (studentLLM.east) -- (ys.west);

    % Cross-Entropy Loss Bracket
    \draw [decorate,decoration={brace,amplitude=5,raise=5pt}, thick]
        (yt.north east) -- (ys.south east) node[midway,right=10pt]{$\ell_\text{CE}(\mathbf{y}_t,\, \mathbf{y}_s)$};

\end{tikzpicture}
\caption{Illustration of \methodname{}. The teacher processes a concatenation (denoted by $\concat$) of demonstrations $\mathbf{X}_d$, consisting of $n$ demonstrations $[\mathbf{x}_1, \mathbf{x}_2, \dots, \mathbf{x}_n]$, and the query $\mathbf{x}_q$. The student, using only the query, fine-tunes its adapter weights to produce outputs $\mathbf{y}_s$ aligned with the teacher's pseudo-labels $\mathbf{y}_t$ by minimizing the cross-entropy loss $\ell_\text{CE}$.}
\label{fig:sticl}
\end{figure}

\section{Method}

\subsection{Disentangling Latent Shifts}

Disentangling in-context knowledge from the query can enhance the efficiency and stability of ICL. Current methods typically achieve disentanglement by modifying the outputs of attention heads or hidden states. The theoretical motivation for this approach stems from previous studies \citep{aizerman-1964-theoretical, irie-etal-2022-dual}, which demonstrate that linear layers optimized via gradient descent can be viewed through the lens of linear attention mechanisms. Specifically, consider a neural network's linear layer characterized by an initial weight matrix \(\mathbf{W}_0 \in \mathbb{R}^{m \times n}\) and an update \(\Delta \mathbf{W} \in \mathbb{R}^{m \times n}\) resulting from backpropagation. Given an input representation \(\mathbf{x} \in \mathbb{R}^{m}\), the linear transformation \(\mathbf{f}: \mathbb{R}^m \to \mathbb{R}^n\) can be expressed succinctly as \(\mathbf{f}(\mathbf{x}) = (\mathbf{W}_0 + \Delta \mathbf{W}) \mathbf{x}\).
Let $\mathbf{x}_i \in \mathbb{R}^{m}$ be a training example and $\mathbf{e}_i \in \mathbb{R}^{n}$ the error signal on $\mathbf{x}_i$ obtained from the gradient of the loss function. During backpropagation, $\Delta \mathbf{W}$ is computed by accumulating the outer products (denoted by $\otimes$) of $N$ training examples $\{\mathbf{x}_1, \mathbf{x}_2, \dots, \mathbf{x}_N\}$ and their error signals $\{\mathbf{e}_1, \mathbf{e}_2, \dots, \mathbf{e}_N\}$, i.e., $\Delta \mathbf{W} = \sum_{i=1}^N \mathbf{e}_i \otimes \mathbf{x}_i$.
The update part of linear layers optimized by gradient descent can be expressed as unnormalized linear dot-product attention \citep{irie-etal-2022-dual}:
\begin{equation}
    \mathbf{f} (\mathbf{x})
    = (\mathbf{W}_0 + \Delta\mathbf{W}) \mathbf{x} 
    = \mathbf{W}_0 \mathbf{x} + \sum_{i=1}^N{(\mathbf{e}_i \otimes \mathbf{x}_i) \mathbf{x}} 
    =  \mathbf{W}_0 \mathbf{x} + \underbrace{\sum_{i=1}^N{\mathbf{e}_i (\mathbf{x}_i^T \mathbf{x}})}_{\text{linear attention}} .
\label{eq:dual}
\end{equation}
In the context of the attention mechanism, this shows that the latent shift $\Delta \mathbf{W} \mathbf{x}$ induced by training examples corresponds directly to the application of linear attention, with error signals $\mathbf{e}_i$ as values, training examples $\mathbf{x}_i$ as keys, and the current input $\mathbf{x}$ as the attention query.

The concept of disentangling the latent shifts described in (\ref{eq:dual}) can be extended to ICL, albeit only under the approximation of linear attention. Let $\mathbf{W}_V$, $\mathbf{W}_K$, and $\mathbf{W}_Q$ denote the weight matrices for values, keys, and queries, respectively. Let $\mathbf{x}_q^{(t)}$ represent the current query token's embedding at step $t$, and $\mathbf{q}^{(t)} = \mathbf{W}_Q \mathbf{x}_q^{(t)}$ is the corresponding attention query vector. The matrix $\mathbf{X}_q = [\mathbf{x}_q^{(1)}, \mathbf{x}_q^{(2)}, \dots, \mathbf{x}_q^{(t-1)}]$ contains all previous query token representations up to $t-1$, and $\mathbf{X}_d$ is the matrix of demonstration token representations. The concatenation $[\mathbf{X}_d; \mathbf{X}_q]$ along the sequence dimension is used to compute the output of a single attention head (AH) at step $t$, expressed as:
\begin{equation}
 \mathbf{f}_\text{AH}(\mathbf{x}_q^{(t)}) = \mathbf{W}_V [\mathbf{X}_d; \mathbf{X}_q] \; \softmax \left( \frac{ \left( \mathbf{W}_K [\mathbf{X}_d; \mathbf{X}_q] \right)^\top \mathbf{q}^{(t)} }{ \sqrt{d} } \right) ,
\end{equation}
where $d$ is the scaling factor (i.e., the dimensionality of the key vectors). By approximating the attention mechanism with linear attention, it becomes possible to disentangle the latent shift of the zero-shot output of an attention head induced by the query from the latent shift induced by the demonstrations \citep{dai-etal-2023-gpt}:
\begin{equation}
\begin{aligned}
    \mathbf{f}_\text{AH}(\mathbf{x}_q^{(t)}) & \approx \mathbf{W}_V [\mathbf{X}_d; \mathbf{X}_q] \left( \mathbf{W}_K [\mathbf{X}_d; \mathbf{X}_q] \right)^\top \mathbf{q}^{(t)} \\
    & = \underbrace{ \mathbf{W}_V \mathbf{X}_q \left( \mathbf{W}_K \mathbf{X}_q \right)^\top }_{\mathbf{W}_{\text{ZS}}} \mathbf{q}^{(t)} + \underbrace{ \mathbf{W}_V \mathbf{X}_d \left( \mathbf{W}_K \mathbf{X}_d \right)^\top }_{\Delta \mathbf{W}_\text{ICL}} \mathbf{q}^{(t)} .
    % & = \left( \mathbf{W}_{\text{ZS}} + \Delta \mathbf{W}_\text{ICL} \right) \mathbf{q}^{(t)} 
\end{aligned}
\label{eq:approx}
\end{equation}
This approximation disentangles the latent shift induced by the demonstrations $\mathbf{X}_d$ from that induced by the query $\mathbf{x}_q^{(t)}$ (see Appendix \ref{app:dual} for a detailed derivation of (\ref{eq:approx})). \textit{The contribution from ICL is captured as a virtual weight update $\Delta \mathbf{W}_\text{ICL}$, corresponding to virtual gradients}, often referred to as ``meta-gradients'' in the literature. The zero-shot latent shift of the query, corresponding to $\mathbf{W}_{\text{ZS}} \mathbf{q}^{(t)}$, reflects the output without demonstrations, providing the initial state. Analogous to $\Delta \mathbf{W} \mathbf{x}$ in (\ref{eq:dual}), the latent shift $\Delta \mathbf{W}_\text{ICL} \mathbf{q}^{(t)}$ reflects the contribution of ICL. Finally, by substituting $\mathbf{h}_\text{ZS} = \mathbf{W}_\text{ZS} \mathbf{q}^{(t)}$ and $\Delta \mathbf{h}_\text{ICL} = \Delta \mathbf{W}_\text{ICL} \mathbf{q}^{(t)}$, we can rewrite the output of an attention head as:
\begin{equation}
    \mathbf{f}_\text{AH}(\mathbf{x}_q^{(t)}) \approx \mathbf{h}_\text{ZS} + \Delta \mathbf{h}_\text{ICL} .
    \label{eq:clean}
\end{equation}

Although transformer-based LLMs employ non-linear attention in practice, many methods \citep{dai-etal-2023-gpt, zhang-etal-2024-batch, todd-etal-2024-function} rely on theoretical assumptions from linear attention, manipulating attention heads or hidden states to approximate latent shift disentanglement. This simplification, however, overlooks key architectural components such as feed-forward layers, activation functions, and residual connections. While effective to a degree, these methods fall short of fully capturing the complex dynamics through which transformers process demonstrations.
In this work, we explore how virtual weight updates can be obtained more directly while preserving the key components of the transformer architecture.

\subsection{Weak Supervision with ICL}

To disentangle the latent shifts induced by in-context demonstrations, we introduce \methodname{}, a method that uses ICL predictions as a form of weak supervision to encode these shifts into reusable adapter parameters \cite{houlsby-etal-2019-parameter, hu-etal-2022-lora}. Instead of focusing narrowly on attention head manipulations, \methodname{} captures the full impact of demonstrations as expressed in the model’s final outputs -- reflecting the combined effects of all components, including attention layers, feed-forward blocks, and residual paths. By aligning with the actual latent shifts induced by ICL, \methodname{} enables the model to embed and reapply in-context knowledge using its full architecture, without relying on repeated prompting.

At the core of \methodname{} is a simple teacher--student framework: the teacher model, $\mathbf{f}_\text{teacher}$, processes both the demonstrations and the query together to generate pseudo-labels without requiring additional labeled data. The student model, $\mathbf{f}_\text{student}$, shares the same architecture as the teacher but includes adapter parameters. Unlike the teacher, the student processes only the query, using the adapter to internalize the knowledge from the demonstrations, as illustrated in Figure \ref{fig:sticl}. Let $\mathbf{x}_q$ denote the query input and $\mathbf{X}_d$ the matrix of demonstration tokens, where each row corresponds to a single demonstration. The empirical loss is defined using the cross-entropy loss function $\ell_\text{CE}$, which aligns the student’s output distribution with the teacher’s full probability distribution over the vocabulary. This enables the student to learn from the full signal provided by the teacher’s output logits. Formally, the empirical loss is
\begin{equation}
\sum_{\mathbf{x}_q \in \mathcal{D}_\text{unlab}} \ell_{\text{CE}}\left( \mathbf{f}_\text{teacher}\left(\left[  \mathbf{X}^*_d; \mathbf{x}_q\right]\right), \mathbf{f}_\text{student}\left(\mathbf{x}_q\right) \right) \text{,}
\label{eq:loss}
\end{equation}
where $\mathcal{D}_\text{unlab}$ is an unlabeled dataset and $\mathbf{X}^*_d$ is a flattened version of $\mathbf{X}_d$.

\methodname{} fundamentally differs from existing approaches, which manipulate attention heads or hidden states at query time, by instead progressively embedding the knowledge from demonstrations into the adapter parameters, denoted $\mathbf{W}_\text{ICL}$. The base LLM parameters, $\mathbf{W}_\text{ZS}$, capture the zero-shot component, while the total model parameters may be represented as $\mathbf{W}_\text{ZS} \oplus \mathbf{W}_\text{ICL}$, where  $\oplus$ denotes the composition of base and adapter parameters.\footnote{Notably, the number of adapter parameters is significantly smaller compared to the base model parameters.} This setup captures the latent shift introduced by the demonstrations through $\mathbf{W}_\text{ICL}$, extending the disentangling process outlined by (\ref{eq:approx}) across the model's entire architecture. The teacher processes the full input sequence $ \left[\mathbf{X}^*_d; \mathbf{x}_q \right] $, while the student processes only the query, applying $ \mathbf{W}_\text{ICL} $ to integrate demonstration knowledge without explicitly processing the demonstrations. Analogously to (\ref{eq:clean}), the latent shift induced by demonstrations can be recovered by decomposing outputs into zero-shot and ICL components. Let $ \mathbf{h}_\text{LLM}(\mathbf{x}_q \mid \mathbf{W}) $ represent the final latent states of an LLM with parameters $ \mathbf{W} $ when processing the input $ \mathbf{x}_q $. The following decomposition holds:
\begin{equation}
   \mathbf{h}_\text{LLM}(\mathbf{x}_q \mid \mathbf{W}_\text{ZS} \oplus \mathbf{W}_\text{ICL}) = \mathbf{h}_\text{LLM}(\mathbf{x}_q \mid \mathbf{W}_\text{ZS}) + \Delta \mathbf{h}_\text{ICL} \text{,}
\label{eq:decomp}
\end{equation}
where $ \Delta \mathbf{h}_\text{ICL} $ encapsulates the latent shift attributable to the demonstrations. \methodname{} encodes the latent shift implicitly within the adapter parameters $\mathbf{W}_\text{ICL}$, which is central to our approach.
However, if necessary, the latent shift can also be explicitly calculated owing to the decomposition in (\ref{eq:decomp}).

\methodname{} achieves stability not only by disentangling demonstration effects, but also through its training dynamics and parametric nature. During training, the same LLM instance serves as both teacher and student across epochs, with the adapter toggled on or off to alternate between roles. Shuffling demonstrations between epochs mitigates order sensitivity, further stabilizing the ICL process. Crucially, \methodname{} leverages its parametric adapter to internalize demonstration-induced shifts, enabling the model to generalize effectively across ID and near-OOD data (see Section~\ref{sec:exp}). This aligns naturally with the W2S generalization paradigm \citep{lang-etal-2024-theoretical}, in which the student is not merely expected to match the teacher but to surpass it.
\methodname{} facilitates this process by compactly encoding latent shifts in a way that supports both pseudo-label correction (refining noisy targets through local consistency) and coverage expansion (generalizing beyond the teacher’s original scope). Together, these effects enable stable extrapolation across the data distribution, aligning with theoretical expectations of W2S generalization \citep{wei-etal-2021-theoretical}.

\section{Experiments}
\label{sec:exp}

\subsection{Experimental Setup}
\label{sec:setup}
We perform our experiments using decoder-only autoregressive language models provided by Hugging Face \citep{wolf-etal-2020-transformers}. Specifically, we employ Llama 3 (8B) \citep{grattafiori-etal-2024-llama3} and Phi 3 (mini 4k) \citep{abdin2024phi3}, along with Llama 2 (7B) \citep{touvron2023llama2} for comparative purposes. Further details about the models are listed in Table \ref{tab:app_models} of the Appendix.

We assess model performance on seven tasks from the GLUE benchmark \citep{wang-etal-2018-glue}, covering single-sequence binary classification (\cola, \sst, \rte), sequence-pair binary classification (\mrpc, \qqp, \qnli), and sequence-pair multi-class classification (\mnli). Evaluation metrics follow established standards: Matthew's correlation for \cola, $F_1$ scores for \mrpc and \qqp, and accuracy for the remaining tasks, with evaluations conducted on the development sets. Additionally, we measure accuracy on selected datasets from the MMLU benchmark \citep{hendrycks-etal-2021-measuring}, specifically ``elementary math'' (\elmath) and ``miscellaneous'' (\textsc{misc}). We further extend our analysis to the ARC-Challenge benchmark~\citep{clark-etal-2018-think} to assess reasoning and multi-hop generalization, which we show in Appendix~\ref{app:additional}.

Predictions are made based on the probability of generating specific verbalizer tokens as the first token output by the models, facilitated by carefully crafted prompts designed explicitly for single-token answers (see Appendix \ref{app:prompt} for detailed templates).

Our experiments compare \methodname with several baselines, including \textbf{Zero-Shot ($\mathbf{0}$-shot)} inference, which generates predictions without demonstrations, \textbf{Standard ICL ($\mathbf{n}$-shot)}, which uses $n$ demonstrations at inference time, and \textbf{Pattern-Based Fine-Tuning (PBFT)} \citep{schick-schutze-2021-exploiting}, which fine-tunes an adapter module on data-specific patterns. We also include two ICL disentanglement methods, \textbf{In-Context Vectors (ICV)} \citep{liu-etal-2023-context}, which leverages hidden-state representations from demonstration examples, and \textbf{Batch-ICL} \citep{zhang-etal-2024-batch}, which aggregates meta-gradients across multiple one-shot runs.
All methods are evaluated using a fixed number of demonstrations, with $n \in \{4, 8, 16, 32\}$. Each experiment is repeated 10 times with different random seeds, resulting in varied demonstration selections across runs. Alongside generalization scores, we report the standard deviation across these runs as an indicator of each method’s stability. Evaluations for GLUE are conducted on the development sets, whereas for the MMLU datasets, we randomly sample 200 instances for evaluation.

We employ three variants of \methodname, each differing in how demonstrations are selected or ordered during training: \textbf{fixed (\wildaf{})} uses a fixed, unchanging set of demonstrations throughout the entire training process, \textbf{shuffle (\wildas{})} uses the same demonstrations throughout training but shuffles their order at the beginning of each epoch, and \textbf{resample (\wildar{})} draws a new set of demonstrations from a larger labeled pool at each epoch.

We utilize LoRA (Low-Rank Adaptation) \citep{hu-etal-2022-lora} for the adapter modules (for both PBFT and \methodname), corresponding to $0.1$--$0.3\%$ of the total parameter count, depending on the model (see Table \ref{tab:app_models} in the Appendix for adapter sizes per model). For each task, we generate pseudo-labels using the teacher model on unlabeled data. Specifically, we use $100$ unlabeled instances ($\mathcal{D}_\text{unlab}$ in (\ref{eq:loss})) for both the GLUE and MMLU benchmarks. Additionally, for GLUE datasets, we experiment with  $200$ and $500$ instances to assess the impact of the amount of unlabeled data on generalization and stability. We experiment only with $100$ unlabeled instances for MMLU datasets due to their limited size. In all of the experiments, we fine-tune the adapter for $10$ epochs. Further experimental details are provided in Appendix \ref{app:exp}.

\subsection{Generalization and Stability}

\begin{table}[]
\centering
\caption{ID generalization scores for the $16$-shot setup and $|\mathcal{D}_\text{unlab}| = 100$. The standard deviations of $10$ runs are shown as subscripts. The highest scores and smallest standard deviations are highlighted in \textbf{bold}, while the second-best scores are \underline{underlined}.}
\begin{adjustbox}{max width=\textwidth}
\begin{tabular}{lllllllllll}
\toprule
& & \multicolumn{7}{c}{\textbf{GLUE}} & \multicolumn{2}{c}{\textbf{MMLU}} \\
\cmidrule(lr){3-9} \cmidrule(lr){10-11}
 & \textbf{Method} & \textbf{\rte} & \textbf{\sst} & \textbf{\qnli} & \textbf{\mnli} & \textbf{\cola} & \textbf{\mrpc} & \textbf{\qqp} & \textbf{\elmath} & \textbf{\misc} \\
\midrule
\multirow{8}{*}{\rotatebox{90}{Llama 3 (8B)}} 
& $0$-shot & $62.3$ & $79.1$ & $64.3$ & $59.9$ & $44.6$ & $63.6$ & $61.1$ & $31.5$ & $62.5$ \\
& $n$-shot & $75.1_{6.5}$ & $93.5_{2.0}$ & $77.0_{5.5}$ & $68.0_{3.0}$ & $58.5_{4.0}$ & $74.0_{2.5}$ & $70.0_{3.0}$ & $43.5_{3.5}$ & $84.0_{4.0}$ \\
& PBFT & $73.2_{3.8}$ & $93.8_{1.5}$ & $77.8_{6.0}$ & $67.4_{3.5}$ & $56.5_{3.0}$ & $72.0_{2.0}$ & $68.0_{2.5}$ & $44.0_{3.8}$ & $83.5_{4.5}$ \\
& ICV & $72.9_{2.7}$ & $92.2_{1.8}$ & $74.5_{6.3}$ & $67.0_{4.2}$ & $57.3_{3.5}$ & $73.4_{2.3}$ & $69.1_{2.8}$ & $41.5_{4.3}$ & $67.0_{4.2}$ \\
& Batch-ICL & $77.8_{4.7}$ & $94.1_{2.2}$ & $78.0_{6.0}$ & $70.9_{3.5}$ & $59.8_{3.7}$ & $75.2_{2.2}$ & $\underline{72.5}_{2.7}$ & $36.2_{4.0}$ & $81.0_{2.5}$ \\
& \wildaf{} & $83.4_{0.3}$ & $95.1_{\textbf{0.6}}$ & $\underline{80.3}_{\textbf{1.4}}$ & $72.1_{2.5}$ & $\underline{63.7}_{\textbf{1.5}}$ & $76.2_{1.8}$ & $71.9_{1.9}$ & $\underline{46.0}_{2.3}$ & $\underline{86.0}_{2.3}$ \\
& \wildas{} & $\underline{86.0}_{\textbf{0.6}}$ & $\textbf{96.1}_{1.2}$ & $\textbf{81.4}_{2.2}$ & $\underline{73.1}_{\textbf{2.0}}$ & $\textbf{64.3}_{2.2}$ & $\textbf{77.7}_{\textbf{1.5}}$ & $\textbf{73.1}_{\textbf{1.8}}$ & $\textbf{49.5}_{\textbf{2.0}}$ & $\textbf{88.0}_{\textbf{2.2}}$ \\
& \wildar{} & $\textbf{86.5}_{3.0}$ & $\underline{95.5}_{0.8}$ & $79.0_{4.3}$ & $\textbf{73.5}_{3.0}$ & $62.5_{2.8}$ & $\underline{76.5}_{1.9}$ & $72.0_{2.2}$ & $44.0_{2.7}$ & $85.5_{3.3}$ \\
\midrule
\multirow{8}{*}{\rotatebox{90}{Phi 3 (mini 4k)}} 
& $0$-shot & $60.6$ & $78.3$ & $61.1$ & $58.1$ & $43.7$ & $63.1$ & $57.8$ & $29.5$ & $52.0$ \\
& $n$-shot & $72.1_{5.2}$ & $90.6_{2.1}$ & $75.6_{3.2}$ & $65.3_{3.1}$ & $55.5_{4.1}$ & $71.1_{2.6}$ & $66.2_{3.7}$ & $37.5_{3.6}$ & $75.5_{4.1}$ \\
& PBFT & $70.6_{4.3}$ & $90.9_{1.9}$ & $73.6_{3.4}$ & $63.6_{3.6}$ & $53.6_{3.1}$ & $69.6_{2.3}$ & $64.6_{2.6}$ & $36.5_{4.1}$ & $73.5_{4.6}$ \\
& ICV & $71.5_{3.1}$ & $89.1_{2.1}$ & $74.3_{3.2}$ & $64.1_{4.1}$ & $54.1_{3.6}$ & $70.8_{2.4}$ & $65.4_{2.9}$ & $36.0_{4.6}$ & $74.0_{4.3}$ \\
& Batch-ICL & $75.3_{4.2}$ & $91.2_{2.6}$ & $76.6_{3.1}$ & $67.1_{3.6}$ & $56.1_{4.1}$ & $72.6_{2.6}$ & $67.3_{2.8}$ & $38.0_{3.9}$ & $76.0_{4.1}$ \\
& \wildaf{} & $\underline{80.4}_{\textbf{1.2}}$ & $92.1_{\textbf{1.6}}$ & $\underline{78.2}_{\textbf{1.3}}$ & $\underline{69.7}_{2.4}$ & $\underline{59.5}_{2.5}$ & $73.5_{2.1}$ & $\underline{68.6}_{2.2}$ & $\underline{40.5}_{3.2}$ & $\underline{77.5}_{3.6}$ \\
& \wildas{} & $\textbf{82.4}_{1.1}$ & $\textbf{93.2}_{\textbf{1.6}}$ & $79.2_{1.4}$ & $\textbf{70.4}_{\textbf{1.1}}$ & $\textbf{60.7}_{\textbf{2.3}}$ & $\textbf{74.1}_{\textbf{1.4}}$ & $\textbf{69.6}_{\textbf{1.9}}$ & $\textbf{41.5}_{\textbf{2.3}}$ & $\textbf{78.0}_{\textbf{3.3}}$ \\
& \wildar{} & $79.0_{1.9}$ & $\underline{92.6}_{2.0}$ & $\textbf{79.6}_{2.9}$ & $68.6_{3.9}$ & $58.6_{2.9}$ & $\underline{73.6}_{2.0}$ & $68.1_{2.3}$ & $39.5_{3.6}$ & $77.0_{3.7}$ \\
\bottomrule
\end{tabular}
\end{adjustbox}
\label{tab:id_gen}
\end{table}

We first evaluate the generalization and stability of \methodname on ID data. Table \ref{tab:id_gen} reports the $16$-shot ID generalization scores along with standard deviations.
Across all datasets and models, \wildas{} consistently achieves the best generalization scores, outperforming standard ICL, PBFT, and the disentanglement methods ICV and Batch-ICL (cf.~Table~\ref{tab:app_id_gen} in the Appendix for results with Llama 2).
Compared to standard ICL, \wildas{} \textit{shows absolute improvements ranging from $2.6\%$ to $11.9\%$ for Llama 3 and $2.5\%$ to $10.3\%$ for Phi 3}, where the differences in scores are statistically significant across all datasets.\footnote{We assess the statistical significance using a two-tailed Wilcoxon signed-rank test ($p < 0.05$), applying the Holm-Bonferroni method for family-wise error rate correction.} Similar patterns hold for $n \in \{4,8,32\}$, where \wildas{} also surpasses standard ICL (cf.~Table \ref{tab:app_nshot} in the Appendix for other $n$-shot setups). Additionally, when a larger set $\mathcal{D}_\text{unlab}$ is used, there is a marginal improvement in scores, while stability improves even further (see Table \ref{tab:app_unlab} in the Appendix).
Notably, the improvements in generalization with \wildas{}, compared to standard ICL (the teacher model in \methodname), provide strong evidence that the student model is exhibiting W2S generalization; we provide a more detailed analysis of this phenomenon in Section \ref{sec:w2s}.
While the \wildaf{} and \wildar{} variants show similar generalization scores to \wildas{}, they typically exhibit higher variance. This makes \wildas{} the preferred choice due to its greater stability with respect to demonstration selection, as it consistently improves upon standard $n$-shot ICL across all datasets and models.
This is supported by the statistically significant differences in standard deviations on all datasets for Llama 3 and on all but \qnli for Phi 3.\footnote{\label{test_footnote} We test for significance using a two-tailed Levene's test ($p < 0.05$) and apply the Holm-Bonferroni method to correct for family-wise error rate.}

Having looked at stability with respect to demonstration selection, we now turn to a more focused evaluation of stability with respect to demonstration ordering. Table \ref{tab:stab} reports the standard deviations across $50$ runs, where the same set of demonstrations is used, but their order is shuffled for each run. Designed to adapt to shuffled demonstrations, \wildas{} \textit{shows the highest stability to demonstration ordering}, as evidenced by the smallest standard deviation. The stability improvements with \wildas{} over standard ICL are statistically significant across all datasets.\footref{test_footnote}

We next assess the capacity of \methodname{} to perform OOD generalization by fine-tuning an adapter on one dataset and then applying the student model to a different dataset within the same task category, simulating a near-OOD scenario with pairs of closely related datasets.
Table \ref{tab:ood_gen} shows the OOD generalization scores for such pairs of datasets in the GLUE benchmark. The results show that \textit{\wildas{} not only outperforms other methods in OOD generalization but also maintains higher stability when adapting to new domains} (cf.~Table \ref{tab:app_ood_gen} in the Appendix for results with other models).

Beyond generalization and stability, we also assess whether the adapters faithfully encode the information contained in the demonstrations. To evaluate this, we encode single demonstrations into the adapters and measure the student model’s ability to reconstruct them using a simple recall task. As detailed in Appendix~\ref{app:faithful}, the adapters achieve consistently high semantic similarity across GLUE datasets, indicating that the demonstration content is reliably preserved. These findings corroborate the effectiveness of \methodname{} in capturing and \textit{storing task-specific information within the adapter weights in a faithful and disentangled manner}.

\begin{table}[]
\centering
\small
\caption{Standard deviations of generalization scores across $50$ runs with varied orderings of $16$ demonstrations. The smallest deviations are in \textbf{bold}, and the second-smallest are \underline{underlined}.}
\begin{adjustbox}{max width=\textwidth}
\begin{tabular}{lllllllllll}
\toprule
& & \multicolumn{7}{c}{\textbf{GLUE}} & \multicolumn{2}{c}{\textbf{MMLU}} \\
\cmidrule(lr){3-9} \cmidrule(lr){10-11}
  \textbf{Model} & \textbf{Method} & \textbf{\rte} & \textbf{\sst} & \textbf{\qnli} & \textbf{\mnli} & \textbf{\cola} & \textbf{\mrpc} & \textbf{\qqp} & \textbf{\elmath} & \textbf{\misc} \\
\midrule
\multirow{7}{*}{\rotatebox{90}{LLama 3 (8B)}} 
& $n$-shot & $4.81$ & $1.62$ & $4.19$ & $2.22$ & $3.04$ & $1.81$ & $2.03$ & $2.52$ & $2.87$ \\
& PBFT & $2.71$ & $1.14$ & $4.53$ & $2.69$ & $2.27$ & $1.57$ & $1.82$ & $2.70$ & $3.22$ \\
& ICV & $2.09$ & $1.23$ & $4.08$ & $2.81$ & $1.95$ & $1.61$ & $2.03$ & $1.96$ & $3.18$ \\
& Batch-ICL & $3.04$ & $1.47$ & $2.89$ & $2.24$ & $2.53$ & $\underline{1.42}$ & $1.74$ & $2.51$ & $2.59$ \\
& \wildaf{} & $\underline{1.32}$ & $\underline{0.72}$ & $\underline{1.53}$ & $\underline{1.83}$ & $\underline{1.76}$ & $1.54$ & $\underline{1.38}$ & $\underline{1.89}$ & $\underline{2.07}$ \\
& \wildas{} & $\textbf{0.22}$ & $\textbf{0.53}$ & $\textbf{1.04}$ & $\textbf{1.21}$ & $\textbf{1.28}$ & $\textbf{0.73}$ & $\textbf{1.14}$ & $\textbf{1.22}$ & $\textbf{0.97}$ \\
& \wildar{} & $2.04$ & $1.34$ & $2.47$ & $2.05$ & $1.85$ & $1.48$ & $1.64$ & $2.03$ & $2.51$ \\
\bottomrule
\end{tabular}
\end{adjustbox}
\label{tab:stab}
\end{table}
\begin{table}[]
\centering
\small
\caption{OOD generalization scores with $16$ shots averaged over 10 runs, with standard deviations shown as subscripts. For each dataset pair, demonstrations are taken from the \textbf{left} dataset, and the model is tested on the \textbf{right} dataset. Columns represent results on the \textbf{right} datasets. The highest scores and lowest standard deviations are in \textbf{bold}, and the second-highest scores are \underline{underlined}. Values in parentheses indicate differences from ID performance for the corresponding target dataset.}
\begin{adjustbox}{max width=\textwidth}
\begin{tabular}{llllll}
\toprule
\textbf{Model} & \textbf{Method} & \textbf{\qnli $\rightarrow$ \rte} & \textbf{\rte $\rightarrow$ \qnli} & \textbf{\qqp $\rightarrow$ \mrpc} & \textbf{\mrpc $\rightarrow$ \qqp} \\
\midrule
\multirow{7}{*}{\rotatebox{90}{Llama 3 (8B)}}
& $n$-shot & $66.3_{2.4} \ (8.8)$ & $69.6_{1.3} \ (7.4)$ & $66.5_{1.9} \ (7.5)$ & $62.2_{2.3} \ (7.8)$ \\
& PBFT & $66.1_{1.5} \ (7.1)$ & $69.1_{1.6}$ \ (8.7) & $67.2_{1.8} \ (4.8)$ & $62.4_{1.2} \ (5.6)$ \\
& ICV & $65.7_{1.2} \ (7.2)$ & $68.7_{2.3} \ (5.8)$ & $67.5_{1.6} \ (5.9)$ & $63.0_{2.1} \ (6.1)$ \\
& Batch-ICL & $65.3_{1.4} \ (12.5)$ & $66.3_{2.5} \ (11.7)$ & $64.9_{2.3} \ (10.3)$ & $62.1_{2.1} \ (10.4)$ \\
& \wildaf{} & $\underline{67.5}_{1.1} \ (15.9)$ & $\underline{70.5}_{1.4} \ (9.8)$ & ${68.5}_{\textbf{1.0}} \ (7.7)$ & ${64.4}_{1.5} \ (7.5)$ \\
& \wildas{} & $\textbf{69.0}_{\textbf{0.5}} \ (17.0)$ & $\textbf{71.3}_{\textbf{0.7}} \ (10.1)$ & $\textbf{69.0}_{2.2} \ (8.7)$ & $\underline{66.4}_{\textbf{1.1}} \ (6.7)$ \\
& \wildar{} & $67.1_{1.7} \ (19.4)$ & $70.0_{1.4} \ (9.0)$ & $68.0_{2.7} \ (8.5)$ & $\textbf{68.3}_{2.0} \ (3.7)$ \\
\bottomrule
\end{tabular}
\end{adjustbox}
\label{tab:ood_gen}
\end{table}

\subsection{Adapter Arithmetic}

To overcome the limitations of context window sizes and efficiently handle extensive demonstration sets in ICL, we employ \textit{adapter arithmetic} within \methodname. 
This is achieved by fine-tuning separate adapters for each demonstration subset, with each adapter encoding the latent shift corresponding to its subset.
Following the approach of \citet{chitale-etal-2023-task}, these adapters are merged by summing their parameters, producing a single adapter that integrates knowledge from all subsets.
Partitioning demonstrations into smaller subsets enables more effective use of the available context window, allowing models to incorporate more demonstrations without exceeding length limits or modifying the base LLM architecture.
Additionally, distributing the prompt across multiple adapters improves GPU utilization by fitting it on a single GPU and reducing memory overhead during inference.

Table \ref{tab:adapters} shows the ID generalization scores of ICV, Batch-ICL, and \methodname in fusing knowledge from multiple demonstration subsets, specifically using $2$, $4$, and $8$ subsets of $16$ demonstrations each.
\wildas{} consistently outperforms baseline methods, highlighting its effectiveness in knowledge fusion across subsets \citep{wan-etal-2024-knowledge}.
Moreover, this form of adapter arithmetic aligns with recent advances in task arithmetic, where merging task-specific parameters promotes generalization across multiple tasks \citep{ilharco-etal-2023-editing, ortiz-jimenez-etal-2023-task}. In our case, \textit{this approach effectively improves generalization and stability when fusing demonstration subsets within the same task.}

\begin{table}[]
\centering
\caption{ID generalization scores of knowledge fusion for Llama 3 (8B). The scores are averaged over $10$ runs with standard deviations shown as subscripts. The table compares the effectiveness of knowledge fusion from $2$, $4$, and $8$ subsets of $16$ demonstrations. The highest scores are in \textbf{bold}.}
\begin{adjustbox}{max width=\textwidth}
\begin{tabular}{cl llllllllll}
\toprule
& & \multicolumn{7}{c}{\textbf{GLUE}} & \multicolumn{2}{c}{\textbf{MMLU}} \\
\cmidrule(lr){3-9} \cmidrule(lr){10-11}
\textbf{Demonstrations} & \textbf{Method} & \textbf{\rte} & \textbf{\sst} & \textbf{\qnli} & \textbf{\mnli} & \textbf{\cola} & \textbf{\mrpc} & \textbf{\qqp} & \textbf{\elmath} & \textbf{\misc} \\
\midrule
\multirow{3}{*}{$\mathbf{2 \times 16}$}
& ICV & $75.2_{4.3}$ & $93.6_{1.9}$ & $77.6_{5.9}$ & $69.2_{3.7}$ & $58.3_{3.5}$ & $74.2_{2.4}$ & $70.6_{2.7}$ & $45.5_{3.7}$ & $72.5_{2.9}$ \\
& Batch-ICL & $80.2_{3.6}$ & $95.3_{1.8}$ & $80.2_{5.8}$ & $72.3_{3.0}$ & $61.2_{3.1}$ & $76.3_{2.0}$ & $72.6_{2.4}$ & $43.5_{2.9}$ & $83.0_{3.6}$ \\
& \wildas{} & $\textbf{87.1}_{1.6}$ & $\textbf{96.4}_{1.3}$ & $\textbf{81.5}_{5.0}$ & $\textbf{75.5}_{2.5}$ & $\textbf{68.4}_{1.8}$ & $\textbf{78.5}_{1.4}$ & $\textbf{74.1}_{1.6}$ & $\textbf{51.5}_{1.6}$ & $\textbf{89.5}_{2.0}$ \\
\midrule
\multirow{3}{*}{$\mathbf{4 \times 16}$}
& ICV & $78.3_{3.6}$ & $94.6_{1.8}$ & $79.3_{5.5}$ & $71.2_{3.1}$ & $60.3_{3.3}$ & $75.6_{2.2}$ & $72.3_{2.4}$ & $47.5_{3.5}$ & $76.5_{3.8}$ \\
& Batch-ICL & $84.4_{3.3}$ & $96.4_{1.5}$ & $82.4_{5.2}$ & $74.3_{2.5}$ & $64.2_{2.8}$ & $78.3_{1.6}$ & $74.3_{2.1}$ & $45.5_{2.6}$ & $84.5_{3.3}$ \\
& \wildas{} & $\textbf{88.4}_{2.3}$ & $\textbf{97.5}_{0.7}$ & $\textbf{83.6}_{4.4}$ & $\textbf{77.3}_{2.2}$ & $\textbf{71.4}_{1.5}$ & $\textbf{79.6}_{0.7}$ & $\textbf{75.2}_{1.3}$ & $\textbf{53.5}_{1.4}$ & $\textbf{91.0}_{1.7}$ \\
\midrule
\multirow{3}{*}{$\mathbf{8 \times 16}$}
& ICV & $81.3_{2.8}$ & $95.6_{1.5}$ & $81.8_{5.0}$ & $73.3_{2.7}$ & $61.3_{2.4}$ & $77.3_{1.7}$ & $73.8_{2.0}$ & $47.5_{2.9}$ & $78.0_{3.5}$ \\
& Batch-ICL & $85.6_{2.5}$ & $96.7_{1.1}$ & $83.8_{4.5}$ & $75.8_{2.1}$ & $65.3_{2.1}$ & $79.8_{1.3}$ & $75.8_{1.8}$ & $45.5_{2.0}$ & $84.0_{2.5}$ \\
& \wildas{} & $\textbf{92.8}_{0.8}$ & $\textbf{98.1}_{0.2}$ & $\textbf{87.9}_{2.5}$ & $\textbf{81.3}_{0.9}$ & $\textbf{74.1}_{0.6}$ & $\textbf{82.8}_{0.4}$ & $\textbf{78.9}_{0.5}$ & $\textbf{57.0}_{0.5}$ & $\textbf{93.0}_{0.7}$ \\
\bottomrule
\end{tabular}
\end{adjustbox}
\label{tab:adapters}
\end{table}

\subsection{Few-Shot \methodname}
\label{sec:fewshot}

The core idea behind \methodname{} is to remove the need for explicit demonstrations by encoding their effect directly into the adapter. However, we also want to verify that standard ICL remains functional when the adapter is active. In particular, we examine whether the latent shift captured by the adapter composes additively with the contextual shift induced by new demonstrations provided during inference.

To this end, we evaluate \wildas{} using Llama 3 (8B) in a mixed few-shot configuration. Specifically, we consider a setup in which the adapter has been fine-tuned to encode $16$ demonstrations and is further provided with an additional $16$ in-context demonstrations during inference. This configuration, denoted as $16/16$, thus combines parameter-based and context-based adaptation. We compare it against several baselines: standard $32$-shot ICL, \wildas{} with $16$ encoded demonstrations and no in-context examples ($0/16$), and \wildas{} with $32$ encoded demonstrations and no in-context examples ($0/32$).

The results, summarized in Table~\ref{tab:wilda_mix}, show that the hybrid $16/16$ configuration outperforms standard $32$-shot ICL across all evaluated datasets, indicating that the adapter and in-context demonstrations reinforce one another rather than interfering. While $16/16$ setup performs slightly below the fully encoded $0/32$ variant, likely because the latter benefits from a dedicated fine-tuning phase, the \textbf{$16/16$} setup demonstrates that \methodname{} can successfully integrate additional ICL prompts on top of a fine-tuned adapter. This finding confirms that standard ICL remains fully compatible with adapter-based tuning, enabling a composition of contextual and parametric adaptation mechanisms within a single framework.

\begin{table}[]
\centering
\caption{Performance comparison of \methodname-S configurations and standard 32-shot ICL using Llama 3 (8B).  
Results are averaged over 10 runs. The notation $n/d$ indicates the number of demonstrations in context ($n$) and encoded in the adapter ($d$).}
\begin{adjustbox}{max width=\textwidth}
\begin{tabular}{lccccccccc}
\toprule
& \multicolumn{7}{c}{\textbf{GLUE}} & \multicolumn{2}{c}{\textbf{MMLU}} \\
\cmidrule(lr){2-8} \cmidrule(lr){9-10}
\textbf{Method} & \textbf{\rte} & \textbf{\sst} & \textbf{\qnli} & \textbf{\mnli} & \textbf{\cola} & \textbf{\mrpc} & \textbf{\qqp} & \textbf{\elmath} & \textbf{\misc} \\
\midrule
$32$-shot ICL & 75.3 & 93.2 & 77.7 & 69.1 & 58.3 & 76.4 & 74.2 & 43.0 & 84.5 \\
\wildas{} (0/32)         & 87.9 & 97.9 & 83.1 & 74.0 & 64.6 & 79.4 & 74.8 & 56.5 & 89.0 \\
\wildas{} (0/16)         & 86.0 & 96.1 & 81.4 & 73.1 & 64.3 & 77.7 & 73.1 & 49.5 & 88.0 \\
\wildas{} (16/16)        & 87.3 & 96.4 & 82.2 & 74.6 & 65.4 & 78.2 & 74.5 & 51.0 & 89.0 \\
\bottomrule
\end{tabular}
\end{adjustbox}
\label{tab:wilda_mix}
\end{table}

\section{Analysis of Weak-to-Strong Generalization}
\label{sec:w2s}

\begin{figure*}[]
    \centering
    \begin{subfigure}{0.49\textwidth}
        \centering
        \includegraphics[width=\linewidth]{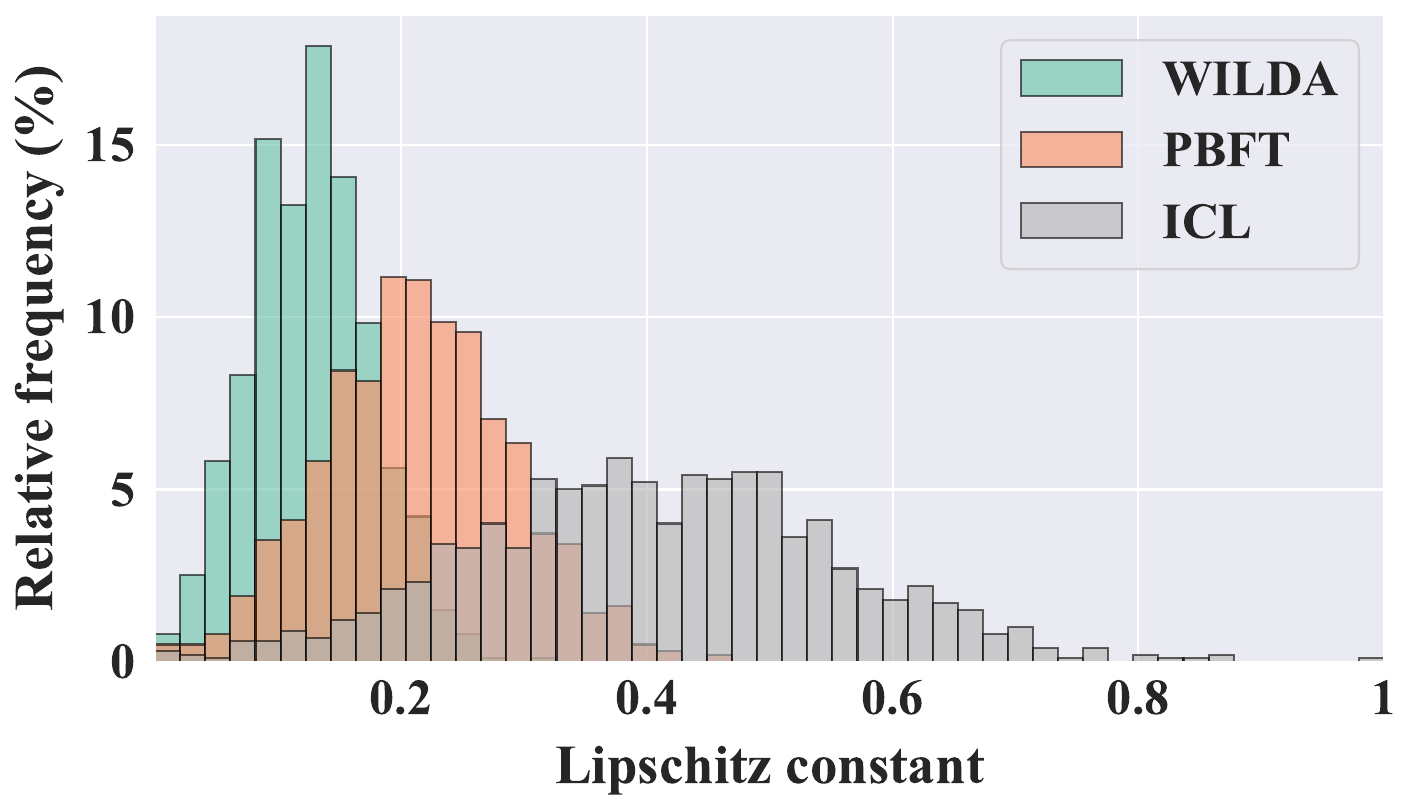}
        \caption{}
        \label{fig:w2s-a}
    \end{subfigure}
    \hfill
    \begin{subfigure}{0.49\textwidth}
        \centering
        \includegraphics[width=\linewidth]{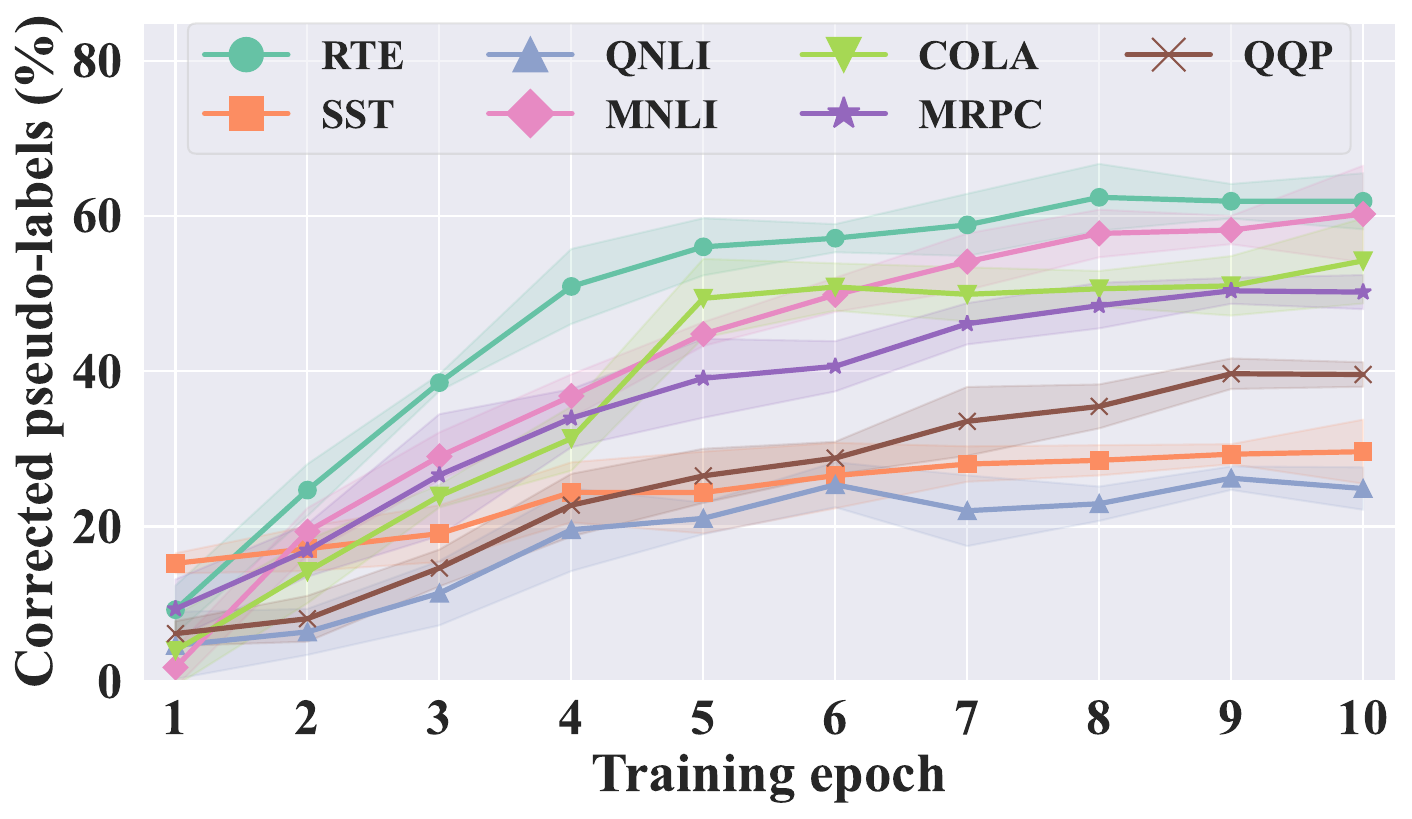}
        \caption{}
        \label{fig:w2s-b}
    \end{subfigure}
    \hfill
    \begin{subfigure}{0.49\textwidth}
        \centering
        \includegraphics[width=\linewidth]{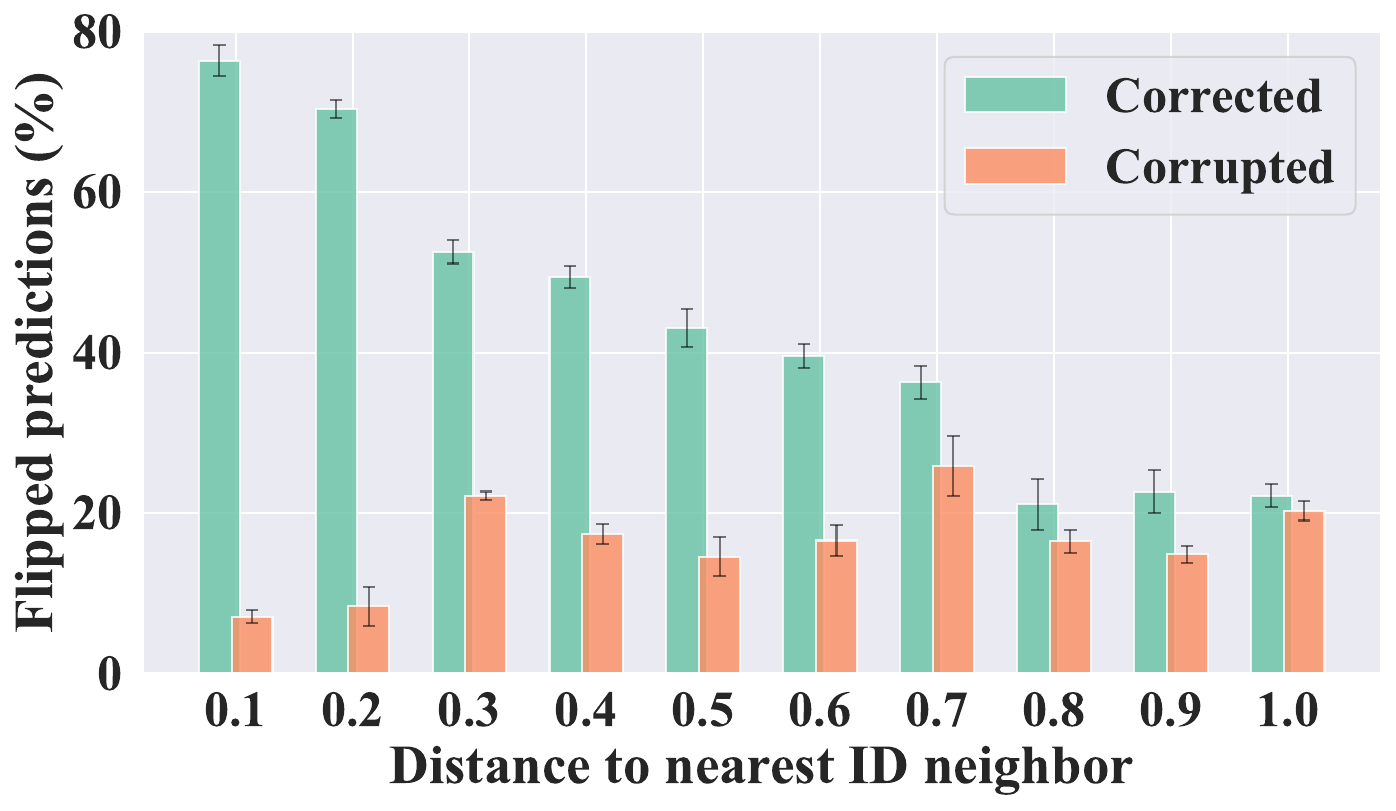}
        \caption{}
        \label{fig:w2s-c}
    \end{subfigure}
    \hfill
    \begin{subfigure}{0.49\textwidth}
        \centering
        \includegraphics[width=\linewidth]{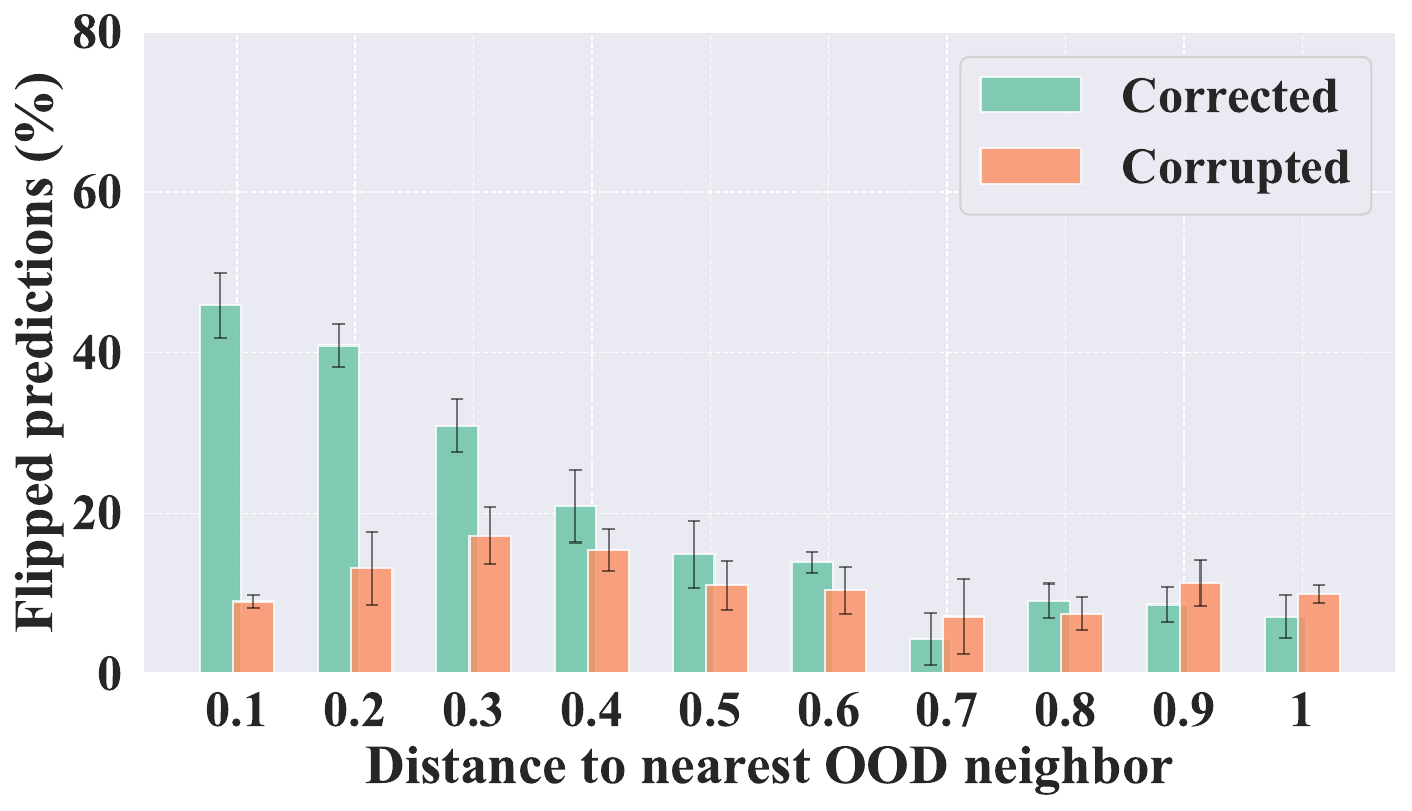}
        \caption{}
        \label{fig:w2s-d}
    \end{subfigure}
     \caption{
        Empirical analysis of \wildas{} on the aggregated GLUE datasets for Llama 3:
        (a) \textbf{Histogram of approximated Lipschitz constants} across datasets, computed as the Frobenius norm of the input--output Jacobian matrix;
        (b) \textbf{Rate of pseudo-label correction} over training epochs (shaded areas indicate the standard deviation over $10$ runs);
        \textbf{Corrected and corrupted prediction rates} for (c) \textbf{ID examples} and (d) \textbf{OOD examples}, based on the normalized Euclidean distance to the nearest correctly pseudo-labeled neighbor. Error bars indicate standard deviation over $10$ runs.
    }
    \label{fig:w2s}
\end{figure*}

Building on the observation that \methodname consistently outperforms its teacher, standard ICL, we hypothesize that W2S generalization may be driving these improvements, where the model's ability to generalize strengthens progressively from weaker signals.
To explore this further, we conduct an empirical analysis of \wildas{} with Llama 3 on aggregated examples from all GLUE datasets, treating them as a single, unified dataset. We focus on the \wildas{} variant due to its consistently strong performance and stability across prior experiments.

%\subsection{Local Consistency}
A crucial prerequisite for successful W2S generalization is the student's ability to maintain stable outputs under small perturbations of the input, i.e., robustness to input variations. A low Lipschitz constant serves as a key indicator of this stability, as it bounds the maximum change in the model output for any change in its input \citep{khromov-and-sidak-2024-some}. However, calculating the exact Lipschitz constant for LLMs is intractable. To approximate it, we leverage the relationship between the Lipschitz constant and the input--output Jacobian matrix of a neural network. Specifically, we compute the Frobenius norm of the Jacobian matrix as a tractable proxy, given its relationship to the spectral norm,
which is a known lower bound for the Lipschitz constant \citep{dherin-etal-2022-neural} (see Appendix \ref{app:lip}). Figure \ref{fig:w2s-a} shows a histogram of the approximated Lipschitz constants (normalized to $[0,1]$) for \methodname, PBFT, and ICL. \methodname exhibits a notably lower Lipschitz constant than PBFT and ICL, which reflects its stronger local consistency.

%\subsection{Pseudo-Label Correction and Coverage Expansion}
The student’s ability to revise the teacher’s predicted labels, known as \textit{pseudo-label correction}, builds directly on local consistency and serves as a key mechanism underlying W2S generalization \citep{lang-etal-2024-theoretical}.
When the model exhibits stable behavior under small input perturbations, as captured by a low Lipschitz constant, it is more likely to propagate corrections reliably across neighboring inputs in representation space. This local consistency forms the foundation for accurate correction of noisy pseudo-labels in high-confidence regions. As corrected labels accumulate, they create a foundation for generalization to nearby, low-confidence examples -- gradually expanding the model’s coverage and facilitating a transition from local consistency to broader generalization.
Figure \ref{fig:w2s-b} shows how the rate of corrected pseudo-labels evolves during training on GLUE datasets. As training progresses, the percentage of corrected pseudo-labels steadily increases, demonstrating \methodname{}'s capacity to exhibit W2S generalization. Notably, the rate of pseudo-label correction plateaus faster for simpler datasets like \sst and \qnli, which have lower linguistic variability.

The mechanism of pseudo-label correction ties into the phenomenon of \textit{coverage expansion}, where the model generalizes beyond the regions covered by pseudo-labels \citep{lang-etal-2024-theoretical}.
We hypothesize that \methodname{}’s generalization ability is supported by coverage expansion, where local corrections gradually influence nearby examples in representation space.
The model’s coverage expands incrementally through a ripple effect, with high-confidence predictions influencing nearby examples while remaining grounded in regions supported by learned corrections.
To understand this dynamic, we analyze which unseen evaluation points are corrected by clustering them based on their proximity to the nearest correctly pseudo-labeled neighbor in $\mathcal{D}_\text{unlab}$. This is quantified by computing the Euclidean distance between the model's representations at the final hidden states, with evaluation points categorized into ten bins based on their normalized distance from the correct neighbor.
Figure \ref{fig:w2s-c} illustrates the rate of prediction flips within these bins, where a flip refers to either correcting an incorrect prediction or corrupting a correct one.
The rate of corrected predictions shows a strong negative association with the distance to the nearest correctly labeled neighbor, as measured by the point biserial correlation coefficient of $-0.968$. In contrast, corrupted predictions are more frequent in regions lacking nearby correct pseudo-labels.
Moreover, coverage expansion also shows its effects on OOD data. Figure~\ref{fig:w2s-d} shows the rate of flipped predictions for OOD data. Although the impact is reduced, a similar correction pattern persists, with a point biserial correlation of $-0.916$.

Overall, the results indicate that \methodname{} does not merely replicate the teacher’s outputs but progressively reorganizes its representation space to reinforce locally consistent decision regions. This smoothing effect allows reliable predictions to extend beyond areas directly supported by demonstrations, consolidating weak contextual supervision into more stable and transferable task knowledge.

\section{Related Work}

% \paragraph{ICL theory.}
Recent perspectives on ICL have shifted from task learning to task identification. \citet{wies-etal-2023-the} argue that ICL works by recognizing latent tasks embedded during pre-training. \citet{hoogland-etal-2024-developmental} build on this, suggesting that ICL unfolds in developmental stages, shedding light on how models adapt to novel contexts. \citet{li-etal-2023-transformers} further empirically show that ICL predictions become more resilient to input perturbations with longer prompts and that training on noisy data enhances stability. Despite these theoretical breakthroughs, ICL remains vulnerable to the selection and ordering of demonstrations \citep{li-etal-2024-debiasing, lu-etal-2021-fantastically}. Moreover, \citet{kossen-etal-2024-incontext} highlight ICL’s biases rooted in pre-training data, revealing that models do not always uniformly leverage in-context information.

% \paragraph{Disentaglement of latent shifts.}
Research into the inner workings of ICL has revealed how transformers process demonstrations to form task representations. \citet{hendel-etal-2023-context} and \citet{liu-etal-2023-context} show that transformers can compress demonstration examples into a task vector, which efficiently directs the model to generate context-appropriate outputs for queries. These task vectors are created during a forward pass, capturing the latent shift induced by the demonstrations. Building on this, \citet{dai-etal-2023-gpt} explore using linear attention to compute virtual gradients, simulating the effect of gradient-based learning within the model. Similarly, \citet{todd-etal-2024-function} use causal mediation analysis to highlight the role of specific attention heads in forming robust task representations in ICL, termed function vectors.

% \paragraph{Weak-to-strong generalization.}
\citet{wei-etal-2021-theoretical} provide a theoretical foundation for W2S generalization, showing that, under the assumption of coverage expansion, models optimized for population-level consistency can achieve high accuracy. \citet{lang-etal-2024-theoretical} further advance this view by formalizing the role of pseudo-label correction, which emerges when a model enforces local consistency during training. Building on these principles, several recent works demonstrate how large language models (LLMs) can leverage their own high-confidence outputs to improve performance. For instance, \citet{huang-etal-2023-large} show that rationale-augmented predictions can guide fine-tuning and enhance reasoning abilities without labeled supervision. Similarly, \citet{qu-etal-2024-recursive} propose recursive introspection for iterative improvement, and \citet{wang-etal-2024-self} introduce self-taught evaluators that enable LLMs to refine their outputs over time.

\section{Conclusion}

To tackle the challenges of stability and long-context handling that arise when processing multiple demonstrations in ICL within LLMs, we introduced \methodname{}, a method that disentangles the latent shifts induced by demonstrations from those of the query, leveraging a teacher--student framework. 
\methodname{} encodes these latent shifts into an adapter module, enabling the student model to handle queries without requiring demonstrations in the input. Moreover, \methodname{} allows efficient handling of large demonstration sets by chunking them into manageable subsets, each processed through separate adapter modules. This not only reduces the instability caused by demonstration selection and ordering but also alleviates the context window limitations inherent in transformer-based models. Additionally, we demonstrated that \methodname{} exhibits weak-to-strong generalization by refining pseudo-labels through progressive corrections, expanding from local consistency to a more comprehensive coverage across the representation space.  Our empirical evaluation confirms these advantages, showing that \methodname{} consistently outperforms traditional ICL methods, significantly improving generalization and stability across diverse natural language understanding datasets.
Ultimately, by treating ICL as weak supervision, \methodname{} enables parametric generalization that can flexibly incorporate contextual adaptation, turning transient in-context behaviors into persistent, reusable parameters and providing a scalable and modular foundation for stable task adaptation in LLMs.

\bibliography{literature}
\bibliographystyle{plainnat}

\newpage
\appendix
\section{Dual Form of ICL}
\label{app:dual}
We offer a detailed derivation of (\ref{eq:approx}), originally introduced by \citep{dai-etal-2023-gpt}, expanding on the key intermediate steps for clarity, which were not explicitly covered in the original work. The goal is to decompose the attention head output into separate components corresponding to the demonstrations and the query, thereby disentangling the latent shifts induced by ICL.

\subsection{Linearized Attention Formulation}

We begin with the approximation of the attention head's output using linear attention:

\begin{equation} \mathbf{f}_\text{AH}(\mathbf{x}_q^{(t)}) \approx \mathbf{W}_V [\mathbf{X}_d ; \mathbf{X}_q] \left( \mathbf{W}_K [\mathbf{X}_d ; \mathbf{X}_q] \right)^\top \mathbf{q}^{(t)}, \label{eq:formulation}
\end{equation}

where: \begin{itemize} \item $\mathbf{W}_V \in \mathbb{R}^{d_h \times d_{\text{model}}}$ is the value weight matrix; \item $\mathbf{W}_K \in \mathbb{R}^{d_h \times d_{\text{model}}}$ is the key weight matrix; \item $\mathbf{X}_d \in \mathbb{R}^{d_{\text{model}} \times N_d}$ is the matrix of demonstration token representations; \item $\mathbf{X}_q \in \mathbb{R}^{d_{\text{model}} \times N_q}$ is the matrix of previous query token representations up to time $t-1$; \item $\mathbf{q}^{(t)} = \mathbf{W}_Q \mathbf{x}_q^{(t)} \in \mathbb{R}^{d_h}$ is the query vector at time $t$, with $\mathbf{W}Q \in \mathbb{R}^{d_h \times d_{\text{model}}}$ being the query weight matrix; \item $[ \mathbf{X}_d ; \mathbf{X}_q ]$ is the concatenation of $\mathbf{X}_d$ and $\mathbf{X}_q$ along the sequence dimension. \end{itemize}

\subsection{Expanding the Concatenated Matrices}

We can expand the concatenated matrices as follows:

\begin{align} \mathbf{W}_V [\mathbf{X}_d ; \mathbf{X}_q] &= [\mathbf{W}_V \mathbf{X}_d ; \mathbf{W}_V \mathbf{X}_q] = [\mathbf{V}_d ; \mathbf{V}_q],
\\ \mathbf{W}_K [\mathbf{X}_d ; \mathbf{X}_q] &= [\mathbf{W}_K \mathbf{X}_d ; \mathbf{W}_K \mathbf{X}_q] = [\mathbf{K}_d ; \mathbf{K}_q], \end{align}

where: \begin{itemize} \item $\mathbf{V}_d = \mathbf{W}_V \mathbf{X}_d$ is the value matrix for the demonstrations; \item $\mathbf{V}_q = \mathbf{W}_V \mathbf{X}_q$ is the value matrix for the previous queries; \item $\mathbf{K}_d = \mathbf{W}_K \mathbf{X}_d$ is the key matrix for the demonstrations; \item $\mathbf{K}_q = \mathbf{W}_K \mathbf{X}_q$ is the key matrix for the previous queries. \end{itemize}

The transpose of the concatenated key matrix is:

\begin{equation} \left( \mathbf{W}_K [\mathbf{X}_d ; \mathbf{X}_q] \right)^\top = \begin{bmatrix} \mathbf{K}_d^\top ; \mathbf{K}_q^\top \end{bmatrix}. \end{equation}

\subsection{Block Matrix Multiplication}

Substituting the expanded forms into Equation (\ref{eq:formulation}) using rules for block matrix multiplication, we have:
\begin{equation} 
\mathbf{f}_\text{AH}(\mathbf{x}_q^{(t)}) \approx \begin{bmatrix} \mathbf{V}_d ; \mathbf{V}_q \end{bmatrix} \begin{bmatrix} \mathbf{K}_d^\top ; \mathbf{K}_q^\top \end{bmatrix} \mathbf{q}^{(t)} = \left( \mathbf{V}_d \mathbf{K}_d^\top + \mathbf{V}_q \mathbf{K}_q^\top \right) \mathbf{q}^{(t)}. 
\label{eq:expanded_forms}
\end{equation}
This separates the contributions from the demonstrations and the query sequences.

\subsection{Component Definitions}

We define:
\begin{align}
\mathbf{W}_{\text{ZS}} &= \mathbf{V}_q \mathbf{K}_q^\top = \mathbf{W}_V \mathbf{X}_q \left( \mathbf{W}_K \mathbf{X}q \right)^\top,
\label{eq:dual_zs}
\\
\Delta \mathbf{W}_\text{ICL} &= \mathbf{V}_d \mathbf{K}_d^\top = \mathbf{W}_V \mathbf{X}_d \left( \mathbf{W}_K \mathbf{X}_d \right)^\top.
\label{eq:dual_icl}
\end{align}

Here: \begin{itemize} \item $\mathbf{W}_{\text{ZS}}$ represents the zero-shot component, capturing the model's behavior based on the query sequence alone; \item $\Delta \mathbf{W}_\text{ICL}$ represents the latent shift induced by the demonstrations, capturing the effect of in-context learning. \end{itemize}

\subsection{Final Decomposition}

Substituting (\ref{eq:dual_zs}) and (\ref{eq:dual_icl}) back into the expression, we obtain:
\begin{equation} 
\mathbf{f}_\text{AH}(\mathbf{x}_q^{(t)}) \approx \left( \mathbf{W}_{\text{ZS}} + \Delta \mathbf{W}_\text{ICL} \right) \mathbf{q}^{(t)} = \mathbf{W}_{\text{ZS}} \mathbf{q}^{(t)} + \Delta \mathbf{W}_\text{ICL} \mathbf{q}^{(t)}.
\end{equation}

The decomposition shows that the attention head output can be viewed as the sum of: \begin{enumerate} \item The \textbf{zero-shot component} ($\mathbf{W}_{\text{ZS}} \mathbf{q}^{(t)}$): the model's output when only the query sequence is considered, without any influence from the demonstrations; \item The \textbf{latent shift due to ICL} ($\Delta \mathbf{W}_\text{ICL} \mathbf{q}^{(t)}$): the additional contribution from the demonstrations, representing the knowledge introduced via in-context learning. \end{enumerate}

This separation aligns with the theoretical motivation to disentangle the latent shifts induced by the demonstrations from those induced by the query, allowing for more efficient and stable processing of queries independently of demonstrations.

\section{Lipschitz Continuity in Neural Networks}
\label{app:lip}
Lipschitz continuity is a fundamental concept in the analysis of neural networks as it provides a bound on how much the output of a function can change with respect to its input. Formally, a function $ f: \mathbb{R}^n \rightarrow \mathbb{R}^m $ is said to be Lipschitz continuous with constant \( L \geq 0 \) if for any two inputs $ \mathbf{x}, \mathbf{x}' \in \mathbb{R}^n $ the following inequality holds:
\[
\|f(\mathbf{x}) - f(\mathbf{x}')\| \leq L \|\mathbf{x} - \mathbf{x}'\|.
\]
This property ensures that the function \( f \) behaves smoothly, meaning small changes in the input lead to small changes in the output, which is crucial for robustness in neural networks, particularly for predictive models \citep{khromov-and-sidak-2024-some}.

\subsection{Relationship Between the Lipschitz Constant and the Jacobian Matrix}
In neural networks, the Lipschitz constant can be bounded by the spectral norm of the Jacobian matrix, which quantifies the sensitivity of a function's output to changes in the input. The Jacobian matrix $ \mathbf{J}_f(\mathbf{x}) \in \mathbb{R}^{m \times n} $ of a function $f$ is defined as the matrix of all partial derivatives:
\[
[\mathbf{J}_f(\mathbf{x})]_{i,j} = \frac{\partial f_i(\mathbf{x})}{\partial x_j}.
\]
The spectral norm of the Jacobian matrix, denoted \( \|\mathbf{J}_f(\mathbf{x})\|_2 \), provides a pointwise lower bound on the global Lipschitz constant \( L \):
\[
\|\mathbf{J}_f(\mathbf{x})\|_2 \leq L, \forall \mathbf{x} \in \mathbb{R}^n.
\]
The spectral norm represents the greatest possible rate of change in the function's output for any input variation. However, calculating the exact spectral norm can be computationally expensive, especially for deep neural networks,  so the Frobenius norm is often used as an efficient alternative.

\subsection{Frobenius Norm as a Surrogate for the Lipschitz Constant}
The Frobenius norm of the Jacobian matrix is often used as a surrogate for estimating the Lipschitz constant to avoid the computational complexity of calculating the spectral norm. The Frobenius norm, denoted $ \|\mathbf{A}\|_F $, is easier to compute and relates to the spectral norm through the following inequality:
\[
\|\mathbf{A}\|_2 \leq \|\mathbf{A}\|_F \leq \sqrt{r} \|\mathbf{A}\|_2,
\]
where $r$ is the rank of the matrix $\mathbf{A}$. While the Frobenius norm generally overestimates the spectral norm, the bounded gap between them implies that reducing the Frobenius norm below its initial value often corresponds to a decrease in the spectral norm as well. This relationship supports its use as a practical proxy for Lipschitz behavior: reductions in the Frobenius norm are generally associated with a lower Lipschitz constant, supported by evidence showing that the Lipschitz constant of neural networks tends to closely track the lower bound defined by the spectral norm \citep{latorre-etal-2020-lipschitz, khromov-and-sidak-2024-some}.

\subsection{Empirical Evaluation of Lipschitz Continuity}
In our experiments, we approximate the Lipschitz constant by computing the Frobenius norm of the input--output Jacobian matrix, where the embeddings are the inputs and the penultimate layer produces the outputs. As shown in Figure \ref{fig:w2s-a}, \methodname demonstrates a significantly lower approximated Lipschitz constant compared to PBFT and ICL. This lower value suggests that \methodname is more robust to input perturbations, which is a critical property for correcting pseudo-labels.

\section{Limitations}
\label{sec:limitations}

\paragraph{Computational cost.}  
\methodname introduces additional computational overhead due to the fine-tuning of adapters. While this fine-tuning is more lightweight compared to full model fine-tuning, it remains more expensive than standard ICL, which avoids weight updates entirely. However, \methodname offsets some of this cost by removing demonstrations from the input during inference. For instance, with Llama 3 (8B) processing $16$ demonstrations from GLUE datasets, inference takes approximately $120$ times longer than a $0$-shot setup (processing only the query). This increased cost scales quadratically with the number of tokens, highlighting the self-attention mechanism as the primary bottleneck when handling $16$ demonstrations. Based on our measurements, fine-tuning with $100$ unlabeled instances and $16$ demonstrations using a single adapter corresponds to the computational cost of approximately $2100$ inferences in a $16$-shot setup. This implies that after about $2100$ inferences, the time spent on fine-tuning is effectively balanced by the reduction in per-inference computational cost.

\paragraph{Applicability.}  
\methodname may be less suitable for scenarios with extremely limited resources, as it relies on access to a supply of unlabeled data. In our experiments with $\{4, 8, 16, 32\}$ demonstrations, we typically used $100$ unlabeled instances, which proved sufficient to achieve strong performance. While unlabeled data is generally easier to acquire than labeled data, there may be scenarios where obtaining even a modest amount of unlabeled data is challenging, potentially limiting the applicability of \methodname.

\paragraph{Large demonstration sets.}  
Although \methodname efficiently encodes demonstrations into adapters to overcome context length limitations, the method has not been extensively tested with very large demonstration sets. From our findings, as the total number of demonstrations increases, using multiple adapters with manageable demonstration sizes tends to be more effective. For instance, we successfully employed $8$ adapters with $16$ demonstrations each (totaling $128$ demonstrations). While this approach theoretically allows for an indefinite increase in the number of demonstrations, its effectiveness with significantly larger sets remains unexplored. Moreover, using additional adapters increases computational costs, introducing a tradeoff between scalability and efficiency.

\section{Additional Results}
\label{app:additional}

\subsection{Supplementary ID and OOD Experiments}

We provide additional experimental results that complement the analyses in the main paper.  
Table~\ref{tab:app_id_gen} presents extended in-domain generalization results under the $16$-shot setup with $|\mathcal{D}_\text{unlab}| = 100$, confirming that \methodname{} consistently outperforms standard ICL and related baselines across GLUE and MMLU benchmarks.  
We further evaluate on the ARC-Challenge benchmark (Table~\ref{tab:arc_challenge}), where \methodname{}-S attains the strongest accuracy, demonstrating that the observed gains extend to more demanding reasoning tasks. 
Table~\ref{tab:app_nshot} examines the effect of the number of demonstrations ($n \in \{4, 8, 32\}$), showing that \methodname{}-S scales predictably with $n$ and remains markedly more data-efficient than standard ICL.  
We also assess the influence of the unlabeled query set size (Table~\ref{tab:app_unlab}), observing steady yet saturating improvements as $|\mathcal{D}_\text{unlab}|$ increases from $200$ to $500$.  
Finally, Table~\ref{tab:app_ood_gen} summarizes out-of-domain generalization results, indicating that \methodname{} preserves strong transfer and stability across heterogeneous evaluation settings.

\begin{table}[]
\centering
\caption{ID generalization scores for the $16$-shot scenario and $|\mathcal{D}_\text{unlab}| = 100$ for Llama 2 (7B). The standard deviations of $10$ runs are shown as subscripts.}
\begin{adjustbox}{max width=\textwidth}
\begin{tabular}{lllllllllll}
\toprule
& & \multicolumn{7}{c}{\textbf{GLUE}} & \multicolumn{2}{c}{\textbf{MMLU}} \\
\cmidrule(lr){3-9} \cmidrule(lr){10-11}
\textbf{Model} & \textbf{Method} & \textbf{\rte} & \textbf{\sst} & \textbf{\qnli} & \textbf{\mnli} & \textbf{\cola} & \textbf{\mrpc} & \textbf{\qqp} & \textbf{\elmath} & \textbf{\misc} \\
\midrule
\multirow{8}{*}{\rotatebox{90}{Llama 2 (7B)}} 
& 0-shot & 57.8 & 75.4 & 59.3 & 55.7 & 40.7 & 59.4 & 58.7 & 29.0 & 59.0 \\
& $n$-shot & 69.2$_{4.3}$ & 89.8$_{2.1}$ & 74.2$_{5.9}$ & 63.3$_{2.8}$ & 54.3$_{3.5}$ & 66.9$_{2.4}$ & 64.7$_{1.5}$ & 37.5$_{4.8}$ & 80.0$_{5.3}$ \\
& PBFT & 69.0$_{2.7}$ & 89.7$_{0.4}$ & 73.3$_{5.0}$ & 64.4$_{4.7}$ & 51.2$_{2.9}$ & 67.9$_{2.0}$ & 64.6$_{1.6}$ & 40.0$_{3.2}$ & 79.5$_{2.1}$ \\
& ICV & 68.0$_{4.6}$ & 87.8$_{2.6}$ & 71.2$_{6.7}$ & 60.9$_{4.0}$ & 53.1$_{2.4}$ & 68.8$_{1.7}$ & 65.0$_{1.9}$ & 39.5$_{2.7}$ & 62.5$_{0.6}$ \\
& Batch-ICL & 75.2$_{0.8}$ & 91.2$_{1.9}$ & 74.0$_{0.8}$ & 66.5$_{3.3}$ & 55.9$_{2.1}$ & 70.3$_{0.8}$ & 69.1$_{1.8}$ & 34.5$_{2.3}$ & 77.0$_{4.1}$ \\
& \wildaf{} & 77.2$_{0.7}$ & 90.2$_{0.7}$ & 76.8$_{4.2}$ & 66.5$_{2.4}$ & 60.1$_{1.2}$ & 71.6$_{0.2}$ & 68.8$_{0.8}$ & 43.0$_{1.6}$ & 82.5$_{2.5}$ \\
& \wildas{} & 81.9$_{2.5}$ & 92.1$_{0.3}$ & 77.3$_{0.9}$ & 70.4$_{1.8}$ & 62.8$_{3.4}$ & 72.3$_{2.6}$ & 68.2$_{0.5}$ & 46.5$_{1.5}$ & 82.5$_{1.7}$ \\
& \wildar{} & 81.1$_{1.9}$ & 93.6$_{2.0}$ & 74.7$_{3.6}$ & 69.6$_{2.9}$ & 57.9$_{2.9}$ & 73.1$_{2.0}$ & 66.8$_{2.3}$ & 41.5$_{2.6}$ & 82.0$_{3.7}$ \\
\bottomrule
\end{tabular}
\end{adjustbox}
\label{tab:app_id_gen}
\end{table}

\begin{table}[]
\centering
\caption{Accuracy on the ARC-Challenge benchmark using Llama 3 (8B) with $|\mathcal{D}_\text{unlab}| = 100$ and $n = 16$ shots. The highest score is shown in \textbf{bold}, and the second-best score is \underline{underlined}.}
\begin{adjustbox}{max width=\textwidth}
\begin{tabular}{lcccccccc}
\toprule
\textbf{Method} & $0$-shot & $n$-shot & PBFT & ICV & Batch-ICL & \wildaf{} & \wildas{} & \wildar{} \\
\midrule
\textbf{Accuracy} & $38.3$ & $54.5$ & $50.8$ & $49.9$ & $48.6$ & $55.7$ & $\mathbf{59.2}$ & \underline{$56.5$} \\
\bottomrule
\end{tabular}
\end{adjustbox}
\label{tab:arc_challenge}
\end{table}

\begin{table}[]
\centering
\caption{ID generalization scores for $n$-shot scenarios ($n = 4, 8, 32$, with $|\mathcal{D}_\text{unlab}| = 100$) for Llama 3 (8B). The standard deviations of $10$ runs are shown as subscripts.}
\begin{adjustbox}{max width=\textwidth}
\begin{tabular}{llllllllllll}
\toprule
& & & \multicolumn{7}{c}{\textbf{GLUE}} & \multicolumn{2}{c}{\textbf{MMLU}} \\
\cmidrule(lr){4-10} \cmidrule(lr){11-12}
\textbf{Model} & \textbf{$n$} & \textbf{Method} & \textbf{\rte} & \textbf{\sst} & \textbf{\qnli} & \textbf{\mnli} & \textbf{\cola} & \textbf{\mrpc} & \textbf{\qqp} & \textbf{\elmath} & \textbf{\misc} \\
\midrule
\multirow{6}{*}{\rotatebox{90}{Llama 3 (8B)}} 
& \multirow{2}{*}{$4$} & $n$-shot & 71.3$_{5.4}$ & 84.5$_{4.4}$ & 70.1$_{2.9}$ & 62.4$_{2.7}$ & 54.6$_{3.5}$ & 69.2$_{4.1}$ & 62.0$_{2.3}$ & 37.0$_{3.9}$ & 76.5$_{2.5}$ \\
& & \wildas{} & 80.3$_{1.5}$ & 90.9$_{0.9}$ & 76.3$_{1.4}$ & 70.1$_{1.8}$ & 61.4$_{2.0}$ & 72.9$_{1.5}$ & 70.3$_{1.2}$ & 43.0$_{1.3}$ & 77.5$_{1.8}$ \\
\cmidrule(lr){2-12}

& \multirow{2}{*}{$8$} & $n$-shot & 72.7$_{2.1}$ & 89.4$_{2.6}$ & 73.5$_{2.5}$ & 64.7$_{3.1}$ & 55.8$_{2.8}$ & 71.2$_{2.4}$ & 64.3$_{2.9}$ & 37.0$_{1.3}$ & 77.5$_{2.1}$ \\
& & \wildas{} & 82.1$_{1.1}$ & 93.2$_{1.0}$ & 78.3$_{1.3}$ & 72.2$_{1.6}$ & 63.7$_{1.8}$ & 73.9$_{1.3}$ & 72.1$_{0.4}$ & 47.5$_{0.5}$ & 84.0$_{1.4}$ \\
\cmidrule(lr){2-12}

& \multirow{2}{*}{$32$} & $n$-shot & 75.3$_{3.2}$ & 93.2$_{1.9}$ & 77.7$_{2.9}$ & 69.1$_{1.9}$ & 58.3$_{1.5}$ & 76.4$_{2.2}$ & 74.2$_{1.9}$ & 43.0$_{1.5}$ & 84.5$_{2.1}$ \\
& & \wildas{} & 87.9$_{0.6}$ & 97.9$_{0.4}$ & 83.1$_{0.9}$ & 74.0$_{1.1}$ & 64.6$_{1.2}$ & 79.4$_{0.6}$ & 74.8$_{1.5}$ & 56.5$_{0.2}$ & 89.0$_{0.4}$ \\

\bottomrule
\end{tabular}
\end{adjustbox}
\label{tab:app_nshot}
\end{table}

\begin{table}[]
\centering
\caption{ID generalization scores of \wildas{} for $n=16$ shots and $|\mathcal{D}_\text{unlab}| = 200, 500$ for Llama 3 (8B). Results are shown for GLUE datasets with $n$-shot and \wildas{} methods. The standard deviations of $10$ runs are shown as subscripts.}
\begin{adjustbox}{max width=\textwidth}
\begin{tabular}{lcllllllll}
\toprule
& & \multicolumn{7}{c}{\textbf{GLUE}} \\
\cmidrule(lr){3-9}
\textbf{Model} & \textbf{$|\mathcal{D}_\text{unlab}|$} &  \textbf{\rte} & \textbf{\sst} & \textbf{\qnli} & \textbf{\mnli} & \textbf{\cola} & \textbf{\mrpc} & \textbf{\qqp} \\
\midrule

\multirow{2}{*}{{Llama 3 (8B)}} 
& 200 & 86.2$_{0.4}$ & 97.2$_{0.4}$ & 81.6$_{1.0}$ & 73.9$_{1.3}$ & 64.7$_{1.1}$ & 78.9$_{0.7}$ & 74.0$_{0.5}$ \\
& 500 & 86.9$_{0.3}$ & 97.1$_{0.5}$ & 81.9$_{0.7}$ & 74.8$_{1.0}$ & 64.6$_{0.8}$ & 81.4$_{0.8}$ & 75.2$_{0.3}$ \\

\bottomrule
\end{tabular}
\end{adjustbox}
\label{tab:app_unlab}
\end{table}
\begin{table}[]
\centering
\small
\caption{OOD generalization scores for Phi 3 and Llama 2 in a $16$-shot scenario with $\mathcal{D}_\text{unlab} = 100$ over $10$ runs with standard deviations shown as subscripts. In each dataset pair, demonstrations are taken from the left dataset, and the model is tested on the right dataset. The columns correspond to the results on the right datasets. }
\begin{adjustbox}{max width=\textwidth}
\begin{tabular}{llcccc}
\toprule
\textbf{Model} & \textbf{Method} & \textbf{\qnli $\rightarrow$ \rte} & \textbf{\rte $\rightarrow$ \qnli} & \textbf{\qqp $\rightarrow$ \mrpc} & \textbf{\mrpc $\rightarrow$ \qqp} \\
\midrule
\multirow{3}{*}{{Phi 3 (mini 4k)}}
& $n$-shot & $64.3_{2.5}$ & $67.2_{1.5}$ & $63.7_{2.3}$ & $59.4_{2.2}$ \\
& PBFT & $64.1_{1.8}$ & $66.9_{1.6}$ & $64.7_{2.0}$ & $60.1_{1.4}$ \\
& \wildas{} & $67.4_{0.6}$ & $69.2_{0.9}$ & $66.3_{2.4}$ & $64.4_{1.3}$ \\
\midrule
\multirow{3}{*}{{Llama 2 (7B)}}
& $n$-shot & $62.9_{2.3}$ & $66.3_{1.2}$ & $64.5_{1.9}$ & $61.1_{2.2}$ \\
& PBFT & $62.8_{1.3}$ & $68.1_{1.4}$ & $65.9_{1.8}$ & $61.3_{1.2}$ \\
& \wildas{} & $64.8_{0.4}$ & $70.3_{0.6}$ & $67.8_{2.1}$ & $65.0_{1.1}$ \\
\bottomrule
\end{tabular}
\end{adjustbox}
\label{tab:app_ood_gen}
\end{table}

\subsection{Task Specificity and Generalization}
\label{app:task-specificity}

LoRA adapters are implemented as additive residuals and do not alter the underlying model parameters. 
During training, we alternate between enabling the adapter (student) and disabling it (teacher), while at inference the adapter can be toggled on or off without modifying the base model. 
To assess whether task-specific adaptation affects general performance, we conducted additional experiments evaluating the model on unrelated, non-overlapping dataset pairs from GLUE. 
In each case, both $16$-shot ICL and \wildas{} were exposed to demonstrations from a different domain.
Results, summarized in Table~\ref{tab:adapter-generalization}, show that performance differences remain within one standard deviation across $10$ random seeds. 
This indicates that \wildas{} does not degrade the model’s general zero-shot capabilities, even when an adapter trained on a specific task remains active during evaluation on a disjoint domain.

\begin{table}[]
\centering
\caption{
Cross-domain impact of \methodname.  
To test whether task-specific adaptation harms general performance, we pair \textit{unrelated} GLUE datasets and evaluate whether demonstrations from one domain deteriorate performance on another.  
Each model variant is exposed to $16$ demonstrations from the unrelated domain.  
\textbf{Demo} denotes the dataset used for constructing demonstrations (i.e., the adapter’s training domain), while \textbf{Eval} indicates the unseen evaluation task.  
We report mean accuracy across $10$ seeds.  
\wildas{} refers to the model with the task-specific adapter \textit{enabled} at inference.}
\label{tab:adapter-generalization}
\begin{tabular}{llccc}
\toprule
\textbf{Demo} & \textbf{Eval} & \textbf{0-shot} & \textbf{16-shot} & \textbf{\wildas{}} \\
\midrule
QNLI & RTE  & $62.3$            & $61.9_{1.2}$ & $62.1_{0.9}$ \\
SST  & COLA & $44.6$           & $44.9_{0.5}$ & $44.7_{0.4}$ \\
MNLI & QQP  & $61.1$            & $60.9_{0.6}$ & $61.0_{0.7}$ \\
MRPC & SST  & $79.1$            & $81.4_{1.4}$ & $81.3_{1.1}$ \\
\bottomrule
\end{tabular}
\end{table}

\subsection{Faithful Encoding and Retrieval of Demonstrations}
\label{app:faithful}

To evaluate whether demonstrations are faithfully encoded and disentangled, we conducted an experiment by encoding a single demonstration into the adapter and assessing the student model's ability to capture this information. Specifically, we utilized $1000$ examples per dataset across the GLUE benchmark using Llama 3 (8B).

For each dataset, the student model was prompted with a simple instruction: ``Repeat the demonstration word for word.'' During the fine-tuning phase, the teacher model processed input examples using the following template: ``Demonstration: \{\textit{demonstration}\}. Answer: (\{\textit{answer}\}).'' The adapter learned to encode demonstration-specific information indirectly by aligning its outputs with the teacher's responses, without explicitly seeing the demonstration itself. After training, the similarity between the student model's response and the original demonstration was computed.
Table \ref{tab:bert_score} shows the average BERTScore similarity \citep{zhang-etal-2020-bertscore} between the original demonstrations and the student's reconstructed response.

\begin{table}[h!]
\centering
\caption{Average BERTScore ($F_1$) similarity across GLUE datasets. Higher scores indicate better fidelity in recalling the encoded demonstration.}
\begin{tabular}{lccccccc}
\toprule
 & \textbf{\rte} & \textbf{\sst} & \textbf{\qnli} & \textbf{\mnli} & \textbf{\cola} & \textbf{\mrpc} & \textbf{\qqp} \\
\midrule
BERTScore & $0.84$ & $0.91$ & $0.80$ & $0.83$ & $0.86$ & $0.82$ & $0.81$ \\
\bottomrule
\end{tabular}
\label{tab:bert_score}
\end{table}

The consistently high BERTScore values across all datasets indicate that the student model can reliably retrieve the encoded demonstration from the adapter. This suggests that \methodname effectively disentangles and stores task-specific information within the adapter's weights. Notably, when compared to standard ICL, \methodname often produced different outputs for certain queries, particularly in instances where it corrected ``corrupted'' labels provided by the teacher. Despite these differences, the student model maintained a high degree of semantic similarity in reproducing the demonstrations. This suggests that the adapter weights capture not only the demonstration itself but also additional latent information that contributes to improved generalization.

We present below a pair of examples from \sst and \rte, chosen to represent reconstructed demonstrations with similarity scores close to the dataset averages.

\begin{example}[\textbf{\sst: Example 1}]
\textbf{Original:}  
\textit{Proves once again he hasn't lost his touch, delivering a superb performance in an admittedly middling film.} \\
\textbf{Answer:} (Positive) \\[0.5em]
\textbf{Reconstructed:}  
\textit{He demonstrates once more that he hasn't missed a beat, delivering a remarkable performance in what is admittedly an average film.} \\
\textbf{Answer:} (Positive)
\end{example}

\begin{example}[\textbf{\sst: Example 2}]
\textbf{Original:}  
\textit{Though many of the actors spark briefly when they first appear, they can't generate enough heat in this cold vacuum of a comedy to ignite a reaction.} \\
\textbf{Answer:} (Negative) \\[0.5em]
\textbf{Reconstructed:}  
\textit{Although some actors manage to show a hint of energy early on, they fail to create any real warmth or spark within this lifeless and chilly comedy.} \\
\textbf{Answer:} (Negative)
\end{example}

\begin{example}[\textbf{\rte: Example 1}]
\textbf{Original:} \\
Premise: \textit{The source added that the investigation proved that the bases of the genocide crime ``were completed with a series of illegal arrests followed in some cases with assassinations or cases of disappearances and were preceded, according to information attached to the file, by cases of torture.''} \\
Hypothesis: \textit{Investigators discovered that a series of illicit arrests were often followed by disappearances or murders and were preceded by torture.} \\
\textbf{Answer:} (True) \\[0.5em]
\textbf{Reconstructed:} \\
Premise: \textit{The investigation confirmed that genocide involved illegal arrests followed by disappearances or murders, often preceded by torture.} \\
Hypothesis: \textit{Investigators found that unlawful arrests frequently resulted in disappearances or murders, often preceded by acts of torture.} \\
\textbf{Answer:} (True)
\end{example}

\begin{example}[\textbf{\rte: Example 2}]
\textbf{Original:} \\
Premise: \textit{American tobacco companies were showing a profit most quarters due to export sales of cigarettes and diversification of products sold, including food.} \\
Hypothesis: \textit{PM often entered markets with both cigarettes and food.} \\
\textbf{Answer:} (False) \\[0.5em]
\textbf{Reconstructed:} \\
Premise: \textit{Profitability was often maintained by American tobacco companies through diversification into food products and successful cigarette exports.} \\
Hypothesis: \textit{Philip Morris International offered food items and cigarettes.} \\
\textbf{Answer:} (False)
\end{example}

\section{Experimental Details}
\label{app:exp}

\subsection{Models}
For all three models -- Llama 3, Llama 2, and Phi 3 -- we utilize the \texttt{bfloat16} half-precision format for parameters. A summary of the models is provided in Table \ref{tab:app_models}.

\subsection{Hyperparameters}
We employ the AdamW optimizer \citep{loshchilov-etal-2017-decoupled} for both PBFT and \methodname variants, with a learning rate of $10^{-4}$. For ICV \citep{liu-etal-2023-context} and Batch-ICL \citep{zhang-etal-2024-batch}, we follow the implementations provided in the original papers and adapt them to our codebase, using their default parameters where specified. In the case of Batch-ICL, we utilize attention heads from the last 20 layers ($k=20$) and fine-tune the model for $10$ epochs.

\paragraph{LoRA adapter configuration.}

\begin{itemize}[leftmargin=1.2em]
    \item \textbf{Rank} ($r = 8$): Dimensionality of the low-rank matrices used to approximate the original weights, controlling parameter efficiency.
    \item \textbf{Scaling factor} ($\alpha = 32$): Multiplier that scales the low-rank updates relative to the frozen base weights.
    \item \textbf{Dropout}: ($0.1$): Regularization applied to the low-rank updates during training.
    \item \textbf{Target modules}: \texttt{q\_proj}, \texttt{k\_proj}, \texttt{v\_proj}, \texttt{o\_proj}, \texttt{gate\_proj}, \texttt{up\_proj}, \texttt{down\_proj}.
\end{itemize}

\subsection{Computing Infrastructure}
We conducted our experiments on \textit{AMD Ryzen Threadripper 3970X 32-Core Processors} and $4 \times$ \textit{NVIDIA GeForce RTX 3090} GPUs with $24$GB of RAM.

\begin{table}[]
\centering
\caption{Summary of the models used in the experiments, including their Hugging Face IDs, parameter counts, context window sizes, training token volumes, and adapter sizes.}
\begin{adjustbox}{max width=\textwidth}
\begin{tabular}{lcccccc}
\toprule
\textbf{Model} & \textbf{Hugging Face ID} & \textbf{Parameters} & \textbf{Context window size} & \textbf{Training tokens} & \textbf{Adapter size} \\
\midrule
Llama 3 & Meta-Llama-3-8B & 8B & 8k & 15T & 21M \\
Llama 2 & Llama-2-7b & 7B & 4k & 2T & 20M \\
Phi 3 & Phi-3-mini-4k-instruct & 3.8B & 4k & 3.3T & 4.5M \\
\bottomrule
\end{tabular}
\end{adjustbox}
\label{tab:app_models}
\end{table}

\section{Prompt Templates}
\label{app:prompt}

\subsection{GLUE Prompt Structure}

\begin{center}
\begin{tcolorbox}[colback=gray!5!white, colframe=gray!75!black, width=\textwidth]
\begin{center}
    \textbf{Generic prompt template for GLUE tasks} \\
\end{center}
\hfill \\
\textbf{Demonstrations:}
\begin{verbatim}
{Sentence 1}
{Sentence 2 (if applicable)}
Answer: ({Correct answer})
\end{verbatim}
\textbf{Query:}
\begin{verbatim}
{Sentence 1}
{Sentence 2 (if applicable)}
Question: {Task-specific question}
Answer: (
\end{verbatim}
\end{tcolorbox}
\end{center}

The prompts for GLUE tasks typically consist of two sentences (or one in certain cases) followed by a task-specific question and the corresponding answer. The model is expected to choose from predefined labels like \textit{Yes/No}, \textit{True/False}, or specific class names based on the dataset. The phrasing of the question preceding each answer in the demonstrations is specific to the task. Below is a list of the questions used for each GLUE dataset. To encourage the model to select from predefined labels, we prepend the phrase ``answer with one word'' before each question, and we append clarifying options such as \textit{Yes or No?} to prompt a more targeted response:

\begin{itemize}
    \item \textbf{RTE:} \{hypothesis\} True or False?
    \item \textbf{SST:} What is the sentiment? Positive or Negative?
    \item \textbf{QNLI:} Does the sentence answer the question? Yes or No?
    \item \textbf{MNLI:} Is the second sentence an Entailment, Contradiction, or Neutral?
    \item \textbf{CoLA:} Is this sentence linguistically acceptable? Yes or No?
    \item \textbf{MRPC:} Do both sentences say the same thing? Yes or No?
    \item \textbf{QQP:} Do both questions ask the same thing? Yes or No?
\end{itemize}

\subsection{MMLU prompt structure}

\begin{center}
\begin{tcolorbox}[colback=gray!5!white, colframe=gray!75!black,  breakable, width=\textwidth]
\begin{center}
    \textbf{Generic prompt template for MMLU sub-datasets} \\
\end{center}
\hfill \\
\textbf{Demonstrations:}
\begin{verbatim}
Question: {Previous Question 1}
Answer choices:
 (A: {Choice A1}),
 (B: {Choice B1}),
 (C: {Choice C1}),
 (D: {Choice D1})
Answer: (Correct Answer 1)

Question: {Previous Question 2}
Answer choices:
(A: {Choice A2}),
(B: {Choice B2}),
(C: {Choice C2}),
(D: {Choice D2})
Answer: (Correct Answer 2)
...
\end{verbatim}
\textbf{Query:}
\begin{verbatim}
Question: {Current Question}
Answer choices:
(A: {Choice A}),
(B: {Choice B}),
(C: {Choice C}),
(D: {Choice D})
Answer: (
\end{verbatim}
\end{tcolorbox}
\end{center}

\begin{center}
\begin{tcolorbox}[colback=gray!5!white, colframe=gray!75!black, breakable, width=\textwidth]
\begin{center}
    \textbf{Example for MMLU \texttt{elementary\_math} (\elmath)} \\
\end{center}
\hfill \\
\textbf{Demonstrations:}
\begin{verbatim}
Question: Ms. Perez drove a total of 40 miles in 5 days.
She drove the same number of miles each day. 
How many miles did Ms. Perez drive each day?
Answer choices: (A: 5), (B: 7), (C: 8), (D: 9)
Answer: (C: 8)

Question: Find the median in the set of data
23, 13, 18, 29, 32, 25.
Answer choices: (A: 18), (B: 24), (C: 25), (D: 29)
Answer: (B: 24)

\end{verbatim}
\textbf{Query:}
\begin{verbatim}
Q: A worker on an assembly line takes 7 hours to produce
22 parts. At that rate how many parts can she produce
in 35 hours?
Answer choices:
(A: 220 parts),
(B: 770 parts),
(C: 4 parts),
(D: 110 parts)
Answer: (
\end{verbatim}
\end{tcolorbox}
\end{center}

\newpage
\section*{NeurIPS Paper Checklist}

\begin{enumerate}

\item {\bf Claims}
    \item[] Question: Do the main claims made in the abstract and introduction accurately reflect the paper's contributions and scope?
    \item[] Answer: \answerYes{} % Replace by \answerYes{}, \answerNo{}, or \answerNA{}.
    \item[] Justification: The abstract and introduction clearly state the goal of disentangling demonstration-induced latent shifts using ICL predictions as weak supervision, and summarize the proposed method (\methodname) and its key contributions, which are consistently supported by the theoretical and empirical results throughout the paper.
    \item[] Guidelines:
    \begin{itemize}
        \item The answer NA means that the abstract and introduction do not include the claims made in the paper.
        \item The abstract and/or introduction should clearly state the claims made, including the contributions made in the paper and important assumptions and limitations. A No or NA answer to this question will not be perceived well by the reviewers. 
        \item The claims made should match theoretical and experimental results, and reflect how much the results can be expected to generalize to other settings. 
        \item It is fine to include aspirational goals as motivation as long as it is clear that these goals are not attained by the paper. 
    \end{itemize}

\item {\bf Limitations}
    \item[] Question: Does the paper discuss the limitations of the work performed by the authors?
    \item[] Answer: \answerYes{}  % Replace by \answerYes{}, \answerNo{}, or \answerNA{}.
    \item[] Justification: We include a discussion of limitations in Appendix \ref{sec:limitations}.
    \item[] Guidelines:
    \begin{itemize}
        \item The answer NA means that the paper has no limitation while the answer No means that the paper has limitations, but those are not discussed in the paper. 
        \item The authors are encouraged to create a separate "Limitations" section in their paper.
        \item The paper should point out any strong assumptions and how robust the results are to violations of these assumptions (e.g., independence assumptions, noiseless settings, model well-specification, asymptotic approximations only holding locally). The authors should reflect on how these assumptions might be violated in practice and what the implications would be.
        \item The authors should reflect on the scope of the claims made, e.g., if the approach was only tested on a few datasets or with a few runs. In general, empirical results often depend on implicit assumptions, which should be articulated.
        \item The authors should reflect on the factors that influence the performance of the approach. For example, a facial recognition algorithm may perform poorly when image resolution is low or images are taken in low lighting. Or a speech-to-text system might not be used reliably to provide closed captions for online lectures because it fails to handle technical jargon.
        \item The authors should discuss the computational efficiency of the proposed algorithms and how they scale with dataset size.
        \item If applicable, the authors should discuss possible limitations of their approach to address problems of privacy and fairness.
        \item While the authors might fear that complete honesty about limitations might be used by reviewers as grounds for rejection, a worse outcome might be that reviewers discover limitations that aren't acknowledged in the paper. The authors should use their best judgment and recognize that individual actions in favor of transparency play an important role in developing norms that preserve the integrity of the community. Reviewers will be specifically instructed to not penalize honesty concerning limitations.
    \end{itemize}

\item {\bf Theory Assumptions and Proofs}
    \item[] Question: For each theoretical result, does the paper provide the full set of assumptions and a complete (and correct) proof?
    \item[] Answer: \answerYes{} % Replace by \answerYes{}, \answerNo{}, or \answerNA{}.
    \item[] Justification: While the paper does not introduce new theorems, it provides complete and correct proofs for existing theoretical results that are necessary to support and motivate the proposed method. All assumptions are clearly stated, and relevant derivations are included in the appendix.
    \item[] Guidelines:
    \begin{itemize}
        \item The answer NA means that the paper does not include theoretical results. 
        \item All the theorems, formulas, and proofs in the paper should be numbered and cross-referenced.
        \item All assumptions should be clearly stated or referenced in the statement of any theorems.
        \item The proofs can either appear in the main paper or the supplemental material, but if they appear in the supplemental material, the authors are encouraged to provide a short proof sketch to provide intuition. 
        \item Inversely, any informal proof provided in the core of the paper should be complemented by formal proofs provided in appendix or supplemental material.
        \item Theorems and Lemmas that the proof relies upon should be properly referenced. 
    \end{itemize}

    \item {\bf Experimental Result Reproducibility}
    \item[] Question: Does the paper fully disclose all the information needed to reproduce the main experimental results of the paper to the extent that it affects the main claims and/or conclusions of the paper (regardless of whether the code and data are provided or not)?
    \item[] Answer: \answerYes{} % Replace by \answerYes{}, \answerNo{}, or \answerNA{}.
    \item[] Justification:  We provide comprehensive implementation details in Section \ref{sec:setup} and in the appendix, including model configurations, training procedures, hyperparameters, and evaluation protocols, sufficient to reproduce all main results.
    \item[] Guidelines:
    \begin{itemize}
        \item The answer NA means that the paper does not include experiments.
        \item If the paper includes experiments, a No answer to this question will not be perceived well by the reviewers: Making the paper reproducible is important, regardless of whether the code and data are provided or not.
        \item If the contribution is a dataset and/or model, the authors should describe the steps taken to make their results reproducible or verifiable. 
        \item Depending on the contribution, reproducibility can be accomplished in various ways. For example, if the contribution is a novel architecture, describing the architecture fully might suffice, or if the contribution is a specific model and empirical evaluation, it may be necessary to either make it possible for others to replicate the model with the same dataset, or provide access to the model. In general. releasing code and data is often one good way to accomplish this, but reproducibility can also be provided via detailed instructions for how to replicate the results, access to a hosted model (e.g., in the case of a large language model), releasing of a model checkpoint, or other means that are appropriate to the research performed.
        \item While NeurIPS does not require releasing code, the conference does require all submissions to provide some reasonable avenue for reproducibility, which may depend on the nature of the contribution. For example
        \begin{enumerate}
            \item If the contribution is primarily a new algorithm, the paper should make it clear how to reproduce that algorithm.
            \item If the contribution is primarily a new model architecture, the paper should describe the architecture clearly and fully.
            \item If the contribution is a new model (e.g., a large language model), then there should either be a way to access this model for reproducing the results or a way to reproduce the model (e.g., with an open-source dataset or instructions for how to construct the dataset).
            \item We recognize that reproducibility may be tricky in some cases, in which case authors are welcome to describe the particular way they provide for reproducibility. In the case of closed-source models, it may be that access to the model is limited in some way (e.g., to registered users), but it should be possible for other researchers to have some path to reproducing or verifying the results.
        \end{enumerate}
    \end{itemize}

\item {\bf Open access to data and code}
    \item[] Question: Does the paper provide open access to the data and code, with sufficient instructions to faithfully reproduce the main experimental results, as described in supplemental material?
    \item[] Answer: \answerYes{} % Replace by \answerYes{}, \answerNo{}, or \answerNA{}.
    \item[] Justification: Documented code for reproducing the main experimental results is provided in the supplemental material at submission time.
    \item[] Guidelines:
    \begin{itemize}
        \item The answer NA means that paper does not include experiments requiring code.
        \item Please see the NeurIPS code and data submission guidelines (\url{https://nips.cc/public/guides/CodeSubmissionPolicy}) for more details.
        \item While we encourage the release of code and data, we understand that this might not be possible, so “No” is an acceptable answer. Papers cannot be rejected simply for not including code, unless this is central to the contribution (e.g., for a new open-source benchmark).
        \item The instructions should contain the exact command and environment needed to run to reproduce the results. See the NeurIPS code and data submission guidelines (\url{https://nips.cc/public/guides/CodeSubmissionPolicy}) for more details.
        \item The authors should provide instructions on data access and preparation, including how to access the raw data, preprocessed data, intermediate data, and generated data, etc.
        \item The authors should provide scripts to reproduce all experimental results for the new proposed method and baselines. If only a subset of experiments are reproducible, they should state which ones are omitted from the script and why.
        \item At submission time, to preserve anonymity, the authors should release anonymized versions (if applicable).
        \item Providing as much information as possible in supplemental material (appended to the paper) is recommended, but including URLs to data and code is permitted.
    \end{itemize}

\item {\bf Experimental Setting/Details}
    \item[] Question: Does the paper specify all the training and test details (e.g., data splits, hyperparameters, how they were chosen, type of optimizer, etc.) necessary to understand the results?
    \item[] Answer: \answerYes{} % Replace by \answerYes{}, \answerNo{}, or \answerNA{}.
    \item[] Justification: All relevant experimental settings, including datasets, splits, model architectures, optimizers, and hyperparameters, are detailed in Section \ref{sec:setup} and in the appendix.
    \item[] Guidelines:
    \begin{itemize}
        \item The answer NA means that the paper does not include experiments.
        \item The experimental setting should be presented in the core of the paper to a level of detail that is necessary to appreciate the results and make sense of them.
        \item The full details can be provided either with the code, in appendix, or as supplemental material.
    \end{itemize}

\item {\bf Experiment Statistical Significance}
    \item[] Question: Does the paper report error bars suitably and correctly defined or other appropriate information about the statistical significance of the experiments?
    \item[] Answer: \answerYes{} % Replace by \answerYes{}, \answerNo{}, or \answerNA{}.
    \item[] Justification: We perform statistical testing of our results and describe the specific tests used in detail. Variability is visualized using standard deviation through error bars and shaded confidence bands, with each plot including an explanation of the statistical measures shown.
    \item[] Guidelines:
    \begin{itemize}
        \item The answer NA means that the paper does not include experiments.
        \item The authors should answer "Yes" if the results are accompanied by error bars, confidence intervals, or statistical significance tests, at least for the experiments that support the main claims of the paper.
        \item The factors of variability that the error bars are capturing should be clearly stated (for example, train/test split, initialization, random drawing of some parameter, or overall run with given experimental conditions).
        \item The method for calculating the error bars should be explained (closed form formula, call to a library function, bootstrap, etc.)
        \item The assumptions made should be given (e.g., Normally distributed errors).
        \item It should be clear whether the error bar is the standard deviation or the standard error of the mean.
        \item It is OK to report 1-sigma error bars, but one should state it. The authors should preferably report a 2-sigma error bar than state that they have a 96\% CI, if the hypothesis of Normality of errors is not verified.
        \item For asymmetric distributions, the authors should be careful not to show in tables or figures symmetric error bars that would yield results that are out of range (e.g. negative error rates).
        \item If error bars are reported in tables or plots, The authors should explain in the text how they were calculated and reference the corresponding figures or tables in the text.
    \end{itemize}

\item {\bf Experiments Compute Resources}
    \item[] Question: For each experiment, does the paper provide sufficient information on the computer resources (type of compute workers, memory, time of execution) needed to reproduce the experiments?
    \item[] Answer: \answerYes{} % Replace by \answerYes{}, \answerNo{}, or \answerNA{}.
    \item[] Justification: We specify the computing infrastructure in the appendix.
    \item[] Guidelines:
    \begin{itemize}
        \item The answer NA means that the paper does not include experiments.
        \item The paper should indicate the type of compute workers CPU or GPU, internal cluster, or cloud provider, including relevant memory and storage.
        \item The paper should provide the amount of compute required for each of the individual experimental runs as well as estimate the total compute. 
        \item The paper should disclose whether the full research project required more compute than the experiments reported in the paper (e.g., preliminary or failed experiments that didn't make it into the paper). 
    \end{itemize}
    
\item {\bf Code Of Ethics}
    \item[] Question: Does the research conducted in the paper conform, in every respect, with the NeurIPS Code of Ethics \url{https://neurips.cc/public/EthicsGuidelines}?
    \item[] Answer: \answerYes{} % Replace by \answerYes{}, \answerNo{}, or \answerNA{}.
    \item[] Justification: he research adheres to the NeurIPS Code of Ethics. It does not involve human subjects, private data, or misuse-prone models.
    \item[] Guidelines:
    \begin{itemize}
        \item The answer NA means that the authors have not reviewed the NeurIPS Code of Ethics.
        \item If the authors answer No, they should explain the special circumstances that require a deviation from the Code of Ethics.
        \item The authors should make sure to preserve anonymity (e.g., if there is a special consideration due to laws or regulations in their jurisdiction).
    \end{itemize}

\item {\bf Broader Impacts}
    \item[] Question: Does the paper discuss both potential positive societal impacts and negative societal impacts of the work performed?
    \item[] Answer: \answerNA{} % Replace by \answerYes{}, \answerNo{}, or \answerNA{}.
    \item[] Justification: The paper introduces a methodological contribution designed for a specific technical context. On its own, the method is not directly applicable in ways that would raise significant societal impact concerns.
    \item[] Guidelines:
    \begin{itemize}
        \item The answer NA means that there is no societal impact of the work performed.
        \item If the authors answer NA or No, they should explain why their work has no societal impact or why the paper does not address societal impact.
        \item Examples of negative societal impacts include potential malicious or unintended uses (e.g., disinformation, generating fake profiles, surveillance), fairness considerations (e.g., deployment of technologies that could make decisions that unfairly impact specific groups), privacy considerations, and security considerations.
        \item The conference expects that many papers will be foundational research and not tied to particular applications, let alone deployments. However, if there is a direct path to any negative applications, the authors should point it out. For example, it is legitimate to point out that an improvement in the quality of generative models could be used to generate deepfakes for disinformation. On the other hand, it is not needed to point out that a generic algorithm for optimizing neural networks could enable people to train models that generate Deepfakes faster.
        \item The authors should consider possible harms that could arise when the technology is being used as intended and functioning correctly, harms that could arise when the technology is being used as intended but gives incorrect results, and harms following from (intentional or unintentional) misuse of the technology.
        \item If there are negative societal impacts, the authors could also discuss possible mitigation strategies (e.g., gated release of models, providing defenses in addition to attacks, mechanisms for monitoring misuse, mechanisms to monitor how a system learns from feedback over time, improving the efficiency and accessibility of ML).
    \end{itemize}
    
\item {\bf Safeguards}
    \item[] Question: Does the paper describe safeguards that have been put in place for responsible release of data or models that have a high risk for misuse (e.g., pretrained language models, image generators, or scraped datasets)?
    \item[] Answer: \answerNA{} % Replace by \answerYes{}, \answerNo{}, or \answerNA{}.
    \item[] Justification: The paper does not release high-risk assets.
    \item[] Guidelines:
    \begin{itemize}
        \item The answer NA means that the paper poses no such risks.
        \item Released models that have a high risk for misuse or dual-use should be released with necessary safeguards to allow for controlled use of the model, for example by requiring that users adhere to usage guidelines or restrictions to access the model or implementing safety filters. 
        \item Datasets that have been scraped from the Internet could pose safety risks. The authors should describe how they avoided releasing unsafe images.
        \item We recognize that providing effective safeguards is challenging, and many papers do not require this, but we encourage authors to take this into account and make a best faith effort.
    \end{itemize}

\item {\bf Licenses for existing assets}
    \item[] Question: Are the creators or original owners of assets (e.g., code, data, models), used in the paper, properly credited and are the license and terms of use explicitly mentioned and properly respected?
    \item[] Answer: \answerYes{} % Replace by \answerYes{}, \answerNo{}, or \answerNA{}.
    \item[] Justification: All datasets and models used (e.g., GLUE, HuggingFace models) are properly cited and used under their respective licenses.
    \item[] Guidelines:
    \begin{itemize}
        \item The answer NA means that the paper does not use existing assets.
        \item The authors should cite the original paper that produced the code package or dataset.
        \item The authors should state which version of the asset is used and, if possible, include a URL.
        \item The name of the license (e.g., CC-BY 4.0) should be included for each asset.
        \item For scraped data from a particular source (e.g., website), the copyright and terms of service of that source should be provided.
        \item If assets are released, the license, copyright information, and terms of use in the package should be provided. For popular datasets, \url{paperswithcode.com/datasets} has curated licenses for some datasets. Their licensing guide can help determine the license of a dataset.
        \item For existing datasets that are re-packaged, both the original license and the license of the derived asset (if it has changed) should be provided.
        \item If this information is not available online, the authors are encouraged to reach out to the asset's creators.
    \end{itemize}

\item {\bf New Assets}
    \item[] Question: Are new assets introduced in the paper well documented and is the documentation provided alongside the assets?
    \item[] Answer: \answerNA{} % Replace by \answerYes{}, \answerNo{}, or \answerNA{}.
    \item[] Justification: We do not introduce new assets in the form of datasets or models. However, we include accompanying code for reproducing our experiments in the supplemental material.
    \item[] Guidelines:
    \begin{itemize}
        \item The answer NA means that the paper does not release new assets.
        \item Researchers should communicate the details of the dataset/code/model as part of their submissions via structured templates. This includes details about training, license, limitations, etc. 
        \item The paper should discuss whether and how consent was obtained from people whose asset is used.
        \item At submission time, remember to anonymize your assets (if applicable). You can either create an anonymized URL or include an anonymized zip file.
    \end{itemize}

\item {\bf Crowdsourcing and Research with Human Subjects}
    \item[] Question: For crowdsourcing experiments and research with human subjects, does the paper include the full text of instructions given to participants and screenshots, if applicable, as well as details about compensation (if any)? 
    \item[] Answer: \answerNA{} % Replace by \answerYes{}, \answerNo{}, or \answerNA{}.
    \item[] Justification: We did not conduct crowdsourcing experiments or research involving human subjects.
    \item[] Guidelines:
    \begin{itemize}
        \item The answer NA means that the paper does not involve crowdsourcing nor research with human subjects.
        \item Including this information in the supplemental material is fine, but if the main contribution of the paper involves human subjects, then as much detail as possible should be included in the main paper. 
        \item According to the NeurIPS Code of Ethics, workers involved in data collection, curation, or other labor should be paid at least the minimum wage in the country of the data collector. 
    \end{itemize}

\item {\bf Institutional Review Board (IRB) Approvals or Equivalent for Research with Human Subjects}
    \item[] Question: Does the paper describe potential risks incurred by study participants, whether such risks were disclosed to the subjects, and whether Institutional Review Board (IRB) approvals (or an equivalent approval/review based on the requirements of your country or institution) were obtained?
    \item[] Answer: \answerNA{} % Replace by \answerYes{}, \answerNo{}, or \answerNA{}.
    \item[] Justification: This research does not involve human subjects and therefore did not require IRB approval.
    \item[] Guidelines:
    \begin{itemize}
        \item The answer NA means that the paper does not involve crowdsourcing nor research with human subjects.
        \item Depending on the country in which research is conducted, IRB approval (or equivalent) may be required for any human subjects research. If you obtained IRB approval, you should clearly state this in the paper. 
        \item We recognize that the procedures for this may vary significantly between institutions and locations, and we expect authors to adhere to the NeurIPS Code of Ethics and the guidelines for their institution. 
        \item For initial submissions, do not include any information that would break anonymity (if applicable), such as the institution conducting the review.
    \end{itemize}

\end{enumerate}

\end{document}